\documentclass[preprint]{elsarticle}

\usepackage{hyperref}
\usepackage[centertags]{amsmath}
\usepackage{amsfonts}
\usepackage{amssymb}
\usepackage{amsmath, bm}
\usepackage[ruled]{algorithm2e}
\usepackage{enumitem}
\usepackage{multicol}
\usepackage{multirow}
\usepackage{makecell}
\usepackage{ragged2e}
\usepackage{changes}
\usepackage{color}
\usepackage{ulem}

\newtheorem{remark}{Remark}
\journal{Journal}
\numberwithin{table}{section}
\usepackage[ruled]{algorithm2e}
\SetKwBlock{GrO}{Grouping optimization:}{end}
\SetKwBlock{PeO}{Perform optimization steps:}{end}

\numberwithin{equation}{section}
\numberwithin{figure}{section}
\begin{document}
% \listoffigures
% \captionsetup[figure]{labelfont={bf},labelformat={default},labelsep=period,name={Figure}}
\begin{frontmatter}
\title{A practical PINN framework for multi-scale problems with multi-magnitude loss terms
%An efficient PINN-based method for multi-scale problems with multi-magnitude loss terms and multi-frequency features
%\tnoteref{tnote1}
}%
% \tnotetext[t1]{This work is supported by the Project of Natural Science Foundation of Shandong Province (no.ZR2021MA092) and the foundation of CAEP(CX20210044).}
% \tnotetext[t2]{The second title footnote which is a longer
% text matter to fill through the whole text width and
% overflow into another line in the footnotes area of the
% first page.}
\author[1,2]{Yong Wang} 
% \ead{wangyong22@gscaep.ac.cn}
\author[1]{Yanzhong Yao\corref{cor1}}
\ead{yao\_yanzhong@iapcm.ac.cn}
\author[2]{Jiawei Guo}

\author[1]{Zhiming Gao}
%\corref{cor1}
%\ead{gao@iapcm.ac.cn}
%\author[1]{S J}

%% Third author's email
%\ead{author3@author.com}
%\author[2]{Given-name4 \snm{Surname4}}
\cortext[cor1]{Corresponding author}
\address[1]{Institute of Applied Physics and Computational
Mathematics, Beijing 100088, China}
\address[2]{Graduate School of China Academy of Engineering Physics, Beijing 100088, China}

\begin{abstract}
For multi-scale problems, the conventional physics-informed neural networks (PINNs) face some challenges in obtaining available predictions. In this paper, based on PINNs, we propose a practical deep learning framework for multi-scale problems by reconstructing the loss function and associating it with special neural network architectures.
New PINN methods derived from the improved PINN framework differ from the conventional PINN method mainly in two aspects. 
First, the new methods use a novel loss function by modifying the standard loss function through a (grouping) regularization strategy. The regularization strategy implements a different power operation on each loss term so that all loss terms composing the loss function are of approximately the same order of magnitude, which makes all loss terms be optimized synchronously during the optimization process.
Second, for the multi-frequency or high-frequency problems, in addition to using the modified loss function, new methods upgrade the neural network architecture from the common fully-connected neural network to special network architectures such as the Fourier feature architecture given in Ref.~\cite{WANG202111} and the integrated architecture developed by us. 
The combination of the above two techniques leads to a significant improvement in the computational accuracy of multi-scale problems.
Several challenging numerical examples demonstrate the effectiveness of the proposed methods. The proposed methods not only significantly outperform the conventional PINN method in terms of computational efficiency and computational accuracy, but also compare favorably with the state-of-the-art methods in the recent literature.
The improved PINN framework facilitates better application of PINNs to multi-scale problems.
The data and code accompanying this paper are available at \href{https://github.com/wangyong1301108/MMPINN}{https://github.com/wangyong1301108/MMPINN}.
\end{abstract}

\end{frontmatter}

% \linenumbers

\section{Introduction}
With the rapid development of artificial intelligence technology and its increasingly widespread application in production and life, deep learning methods represented by PINN (Physics-informed neural networks) \cite{w712178,RAISSI1,Karniadakis,Lu} have gradually become a popular class of methods for solving partial differential equations (PDEs).
Compared with traditional numerical methods, 
%for solving PDEs,
such as finite difference, finite element and finite volume methods, the PINN methods have the following obvious advantages:
they are mesh-free, thus avoiding the hassles of mesh generation on complex regions and the construction of high-precision discrete schemes on meshes of poor geometric quality; 
they are applicable to high-dimensional problems, and the construction and implementation of the methods are dimensionality independent; 
they produce a prediction function over the entire computational domain, rather than a discrete solution on the meshes as in mesh-based methods. 
The PINN methods have received much attention for their ability to integrate physical laws and labeled data into loss functions, and have led to many notable results in computational science and engineering
\cite{Hiddenfluidmechanics123,Surrogatemodeling,ZHANG2022104243}. 
More and more researchers have realized the immense potential of deep learning algorithms.

For multi-scale problems, considering their importance in practical applications \cite{KANG2023124305,WANG2023107467,MRE22},
many researchers have devoted themselves to studying the deep learning methods. 
In Ref.~\cite{CiCP-28-1970}, based on the idea of radial scaling in the frequency domain and activation functions with compact support, Liu et al.~propose Multi-scale Deep Neural Networks (MscaleDNNs) to solve elliptic PDEs with rich frequency contents. 
In Ref.~\cite{LI2023112242}, combining traditional numerical analysis and the MscaleDNNs algorithm, Li et al.~construct $\mathrm{SD}^2\mathrm{NN}$ based on subspace decomposition to capture the smoothed and oscillatory parts of the multi-scale solution for
elliptic type multi-scale PDEs. 
In Ref.~\cite{shijinmazhengwukeke},
Jin et al.~devise an asymptotic-preserving neural networks to solve the multi-scale uncertain linear transport equation with diffusive scaling, overcoming the computational challenges of the curse of dimensionality and the multi-scale problems.
The above works have obtained distinctive research results for specific models such as elliptic equations, transport equations, etc., which provide some new ideas for solving multi-scale problems using deep learning method. However, it is still a topic for further research to construct a general deep learning framework for multi-scale problems based on the analysis of their common characteristics, so as to solve multi-scale models with high accuracy under a unified framework.

The motivation of this paper is to construct a practical deep learning framework based on PINNs for multi-scale problems. This is undoubtedly a challenging task.
In Refs.~\cite{WANG202111, gradientpathologies,WANG200022110768}, Wang et al.~mention that the conventional PINN method
%presented in Ref.~\cite{RAISSI1} 
has some weaknesses for multi-scale problems, which is proved by a large number of numerical experiments.
Through investigation, we believe that the main reasons why the conventional PINN method is difficult to train multi-scale problems are as follows: For most multi-scale problems, there is usually a very large order of magnitude difference between the supervised term and the residual term of the loss function. When the loss function is optimized, the loss term with a small order of magnitude is often not optimized, and even gets worse. Wang et al.~also find this phenomenon.  In Ref.~\cite{gradientpathologies}, they point out that due to the gradient pathologies in PINNs, there will be a dominance of PDE residuals in the optimization process, which will lead to the training being heavily biased towards the residual term, and to larger errors in the fitting of the supervised term.
In addition, for the multi-scale problems with multi-frequency or high-frequency features, solving them by the conventional PINN method not only encounters the mentioned problem of magnitude differences, but also struggles to learn high-frequency functions due to spectral bias \cite{pmlr-v97-rahaman19a,Moseley},  %caused by the deep neural network architecture
finally resulting in inaccurate prediction results.

According to the above analysis, one approach to address the multi-scale problems is to mitigate the adverse effects caused by the large magnitude difference between the loss terms.
Adjusting the weights of the loss terms is the simplest way to balance the magnitude difference,
%balance the difference in magnitude between the loss terms, 
and much work has been done in this area.
In Ref.~\cite{METHODADUALDIMERMETHOD}, based on the dual-dimer method, Liu et al.~present a new neural network with minimax architecture that can systematically adjust the weights of different losses.
In Ref.~\cite{Self-adaptivepinnjcp},  McClenny et al.~give a Self-Adaptive Physics-informed Neural Networks (SA-PINN) using a soft attention mechanism, where different weights are set for each point of the residual and supervisory terms. 
%During the training of the PINN, not only the hyperparameters of the neural networks are updated, but also the weights of each point.  
%In Ref.~\cite{dasapinnpinnpinn}, Zhang et al.~propose differentiable adversarial self-adaptive weighting physics-informed neural networks to automatically optimize the point-wise loss weights.
In addition, to avoid the adverse effects caused by the magnitude difference, a possible way is the \textit{hard constraint approach} ~\cite{w712178,Moseley}, which strictly enforces the initial and boundary conditions by using the neural network as part of a solution ansatz.
%In addition, to avoid the adverse effects caused by the large  difference between the loss terms, a possible way is the "hard constraint approach \cite{w712178,Moseley}", which strictly enforces the initial and boundary conditions by using the neural network as part of a solution ansatz.

%In addition, we could replace deep neural networks with other neural network architectures to improve the performance  of the conventional PINN method for multi-scale problems.}
Another approach to improve the performance  of the conventional PINN method for multi-scale problems  is to replace deep neural networks with other neural network architectures.
In Ref.~\cite{WANG202111},
through the lens of neural tangent kernel theory, Wang et al.~construct novel architectures that effectively tackle the problems with multi-scale behavior.
In Ref.~\cite{gradientpathologies}, the authors construct an improved fully-connected neural networks that could reduce the stiffness of the gradient flows, and provide more accurate results for some multi-scale problems.
In Refs.~\cite{CPINNCPINNCPINNCPINNCPINN} and \cite{CiCPXPINN}, 
Jagtap et al.~devise conservative PINN and extended PINN  via domain decomposition, 
which are particularly effective for multi-scale problems.
%But, arbitrarily choosing weights by trial and error is extremely tedious and may not produce satisfactory results \cite{gradientpathologies}, and 
However,
%for the multi-scale problems with multi-magnitude loss terms, 
employing the appropriate architecture solely does not ensure accurate prediction for multi-scale problems \cite{WANG202111}.
%\wy{So it is necessary to further explore the more efficient method to solve the multi-scale problems.}
%and combine it with the existing improved architecture for some specific multi-scale problems \cite{WANG202111}.}

In this paper, by constructing novel loss functions, we eliminate the adverse effects caused by the large difference in magnitude between the loss terms. Furthermore, by introducing and developing special neural network architectures, we find an effective way to address the multi-frequency problems. By combining the two approaches, we present a practical PINN framework which has the ability to well handle the multi-scale problems with multi-magnitude loss terms.
The rest of this paper is organized as follows:
 In Section 2, the conventional PINN framework for general partial differential equations is introduced. 
In Section 3, we propose an improved PINN framework for multi-scale models, which is named as MMPINN.
In Section 4, some numerical examples are discussed to demonstrate the performance of the new PINN framework.  In the last section, we give a conclusion of our work.

\section{Conventional PINN framework}\label{sec:std-pinn}
The conventional PINN method presented in Ref.~\cite{RAISSI1} utilizes deep neural networks to solve PDEs. The prediction function is obtained by optimizing the loss function through some optimization algorithms. The loss function consists of the supervised term and the residual term. The supervised term reflects the degree of approximation of the definite condition. The residual term, obtained by the automatic differentiation technique, reflects the degree of approximation of the governing equation.
The PINN methods are applicable to a wide range of types of PDEs, including hydrodynamic equations, heat conduction equations, electromagnetism equations, and so on.

Suppose the PDE has the following general forms:
\begin{equation}\label{pde function}
\begin{cases}
\mathcal{P}(u)=f(x), \quad x\in \Omega \subset \mathbb{R}^{d}, \\
\mathcal{B}(u)=g(x), \enspace \quad   x 
\in 
%\Gamma (\textbf{X}) \subset 
\partial\Omega, 
\end{cases}
\end{equation}
where $\mathcal{P}$ denotes the differential operator, $u=u(x)$ is the solution function on the $d$-dimensional domain $\Omega$,
$f(x)$ is the source term, $\mathcal{B}(u)$ represents the definite conditions, including the initial conditions (I.C.) and boundary conditions (B.C.) such as Dirichlet, Neumann, or Robin conditions. 
Note that, the time variable $t$ can be considered a component of $x$ in some cases \cite{Lu} .

The PINN method produces the prediction function, denoted by $u_\theta(x)$, as an approximation of the solution of Eq.~\eqref{pde function}, where $\theta$ is a set containing the weights and biases of the neural network, which are also the parameters we want to obtain by optimizing the loss function.
For the conventional PINN method, the loss function consists of two parts as follows:
\begin{equation}\label{loss_std}
\mathcal{L}(\theta;\Sigma)=w_s\mathcal{L}_s(\theta;\tau_b)+w_r\mathcal{L}_r(\theta;\tau_r),
\end{equation}
where $w_s$ and $w_r$ are the weights for the two parts of the loss function.
In Eq.~\eqref{loss_std}, the \emph{supervised loss term} and the \emph{residual loss term} are
\begin{align}
&\mathcal{L}_{s}(\theta;\tau_{s}) = \frac{1}{N_s}\sum_{i=1}^{N_s}\left|\mathcal{B}(u_\theta(x_i))-g(x_i)\right|^2,\\
&\mathcal{L}_{r}(\theta;\tau_{r})=\frac{1}{N_r}\sum_{i=1}^{N_{r}}\left| \mathcal{P}(u_\theta(x_i))-f(x_i)
\right|^2.  \label{Lr_loss}
\end{align}
$\Sigma = \{\tau_s,\tau_r\}$ denotes the set of training points, where
$\tau_s=\{x_i|_{i=1}^{N_s}, x_i\in\partial\Omega\}$,  and $\tau_{r}=\{x_i|_{i=1}^{N_{r}}, x_i \in\Omega\}$.
$N_s$ and $N_{r}$ denote the number of boundary sampling points on $\partial\Omega$ and the number of inner sampling points in $\Omega$, respectively.

Suppose 
\begin{equation}\label{MIN}
\bar{\theta} = \arg\min_\theta\mathcal{L}(\theta;\Sigma),
\end{equation}
and then $u_{\bar{\theta}}(x)$ is the approximation of the unknown function $u$.
$\bar{\theta}$ is obtained by using certain optimization algorithms to optimize the function $\mathcal{L}(\theta;\Sigma)$. The most commonly used optimization algorithms are  Adam \cite{kingma2017adam} and L-BFGS \cite{LLLBFGS}.

Since the heat equation is a class of equations with a very wide range of applications, the specific form of the loss function for this class of equations is given below as an example.
The heat conduction problem can be formulated by the following equation:
\begin{equation}\label{heatEq}
\begin{cases}
\displaystyle\frac{\partial u}{\partial t}-a^2 \Delta u=Q(x),&x\in \Omega \subset \mathbb{R}^{d},t\in(0,T],\\
u(x,0)=\phi(x),&x\in \Omega \subset \mathbb{R}^{d},\\
u(x,t)=\psi(x, t),&x\in \partial \Omega, t\in(0,T].\\
\end{cases}
\end{equation}
The loss function for Eq.~\eqref{heatEq} is defined as follows:
\begin{align}\label{heatEloss_function}
&\mathcal{L}(\theta;\Sigma)=w_s\mathcal{L}_s(\theta;\tau_{s})+w_r\mathcal{L}_r(\theta;\tau_{r}),\\
&\mathcal{L}_s(\theta;\tau_{s})=\frac{1}{N_0} \sum_{i=1}^{N_0}\left| u_\theta \left(x_i,0\right)-\phi(x_i)  \right|^2 + \frac{1}{N_b} \sum_{i=1}^{N_b}\left|  u_\theta \left(x_i,t_i\right)    - \psi(x_i,t_i)  \right|^2 , \\
&\mathcal{L}_r(\theta;\tau_{r})=\frac{1}{N_r} \sum_{i=1}^{N_r}\left| \frac{\displaystyle{\partial u_{\theta}}}{\displaystyle{\partial t}}(x_i,t_i) -a^2 \Delta u_\theta(x_i,t_i) - Q(x_i,t_i) \right|^2,
\end{align}
where $N_0$ is the number of sampling points for the I.C., $N_b$ is the number of sampling points on the boundary $\partial\Omega$, and $N_r$ is the number of residual points sampled in the computation domain $\Omega$.

\section{An improved PINN framework for multi-scale problems}

In this section, we first make an analysis of the difficulties in solving the multi-scale problems by the conventional PINN method, and then discuss some strategies to overcome these difficulties. Finally, for the multi-scale problems containing multi-magnitude loss terms and multi-frequency features, we apply all these strategies to present an improved PINN framework, which we call the \textbf{MMPINN} framework.  

%For the multi-scale problems, it is difficult to obtain available predictions using the conventional PINN method with the loss function \eqref{loss_std},
% or~\eqref{heatEloss_function},
%for which an improved PINN framework is given here.
%and further propose a solution for the multi-scale problems between different subdomains.
%\wy{In addition, we introduce different neural network architectures for the multi-scale problems, and further propose a new neural network architecture to  mitigate spectral bias and stiffness of gradient flow. 

\subsection{Problems in solving multi-scale models by conventional PINN methods}\label{pro22}
Many problems in practical engineering applications have multi-scale features. For example, the radiation heat conduction problem in inertial confinement fusion is a typical multi-scale problem, where the temperature of the matter can vary from room temperature to millions or even tens of millions of degrees Celsius \cite{MRE1}. There are also some classical multi-scale problems in hydrodynamics, such as the fluid in a large river consisting of both low-frequency, large-amplitude waves and high-frequency, small-amplitude eddies.

For the model with multi-scale feature, the conventional PINN method has some issues in solving them. One of the most prominent issue is that, if there is a large order of magnitude difference between the loss terms of the loss function, a bad phenomenon like the following will occur in the optimization process:

If 
\begin{align*}
\mathcal{L}_r(\theta;\tau_{r}) \gg \mathcal{L}_s(\theta;\tau_{s}),
\end{align*}
 and the optimization objective is 
 \begin{align*}
 \mathcal{L}(\theta;\Sigma) = \mathcal{L}_s(\theta;\tau_{s}) + \mathcal{L}_r(\theta;\tau_{r}),
 \end{align*}
to make the value of $\mathcal{L}_r(\theta;\tau_{r})$ decrease significantly, the value of $\mathcal{L}_s(\theta;\tau_{s})$ will increase instead of decreasing at the beginning of the optimization process until their magnitudes are not particularly different.
The risk associated with this phenomenon is that since the loss function $\mathcal{L}(\theta;\Sigma)$ is generally not convex, the optimization algorithm may fall into a local minimum, at which point the value of $\mathcal{L}_s(\theta;\tau_{s})$ may still be large, indicating that the resulting network prediction function does not satisfy the initial boundary condition.
We know that the functions that satisfy the equation can be a family of functions. If $u(x, t)$ satisfies the heat equation 
\begin{align*}
   u_t-u_{xx}=f(x,y),
\end{align*}
then $u+ax+b$ all satisfy the equation. The initial boundary condition, also known as the definite condition, can be used to uniquely determine the function that satisfies the equation. Therefore, failure to satisfy the initial boundary condition means that the prediction obtained by the neural network is a function that deviates from the true solution.

In addition to the fact that $\mathcal{L}_r(\theta;\tau_{r})$ and $\mathcal{L}_s(\theta;\tau_{s})$ can have large magnitude differences, $\mathcal{L}_r(\theta;\tau_{r})$ can also have large magnitude differences for different computational subdomains. In this case, a similarly bad situation can occur, where the optimization of subdomains with large magnitudes also comes at the expense of the optimization of subdomains with small magnitudes. The final prediction results are often better for subdomains with large magnitudes, while subdomains with small magnitudes have large relative errors or even distortions. %This is like trying to take a photo that includes both an elephant and an ant; by focusing excessively on the elephant, the ant becomes unrecognizable due to significant relative error.\jw{Is this description suitable here?}

When optimizing an objective function that contains loss terms of different orders of magnitude, it is certainly possible to improve the optimization effect of the overall objective by choosing a powerful optimization algorithm or adjusting the parameters of the optimization algorithm, and further to make each loss term as small as possible, but then this is usually difficult to do. 
A more scientific approach is to design reasonable regularization methods to deal with each loss term, translate them to a certain range, so as to  mitigate the difference in magnitude between loss terms. This enables the importance of each loss term to be balanced so that they are optimized simultaneously in the optimization process, which is the main motivation of this paper.

Another important issue of the conventional PINN method is that for the high-frequency or multi-frequency problems, the expressiveness of conventional deep neural networks is limited due to spectral bias \cite{WANG200022110768}. To achieve better prediction results, it is necessary to use a neural network architecture adapted to the problem, such as the multi-scale Fourier feature network architecture in Ref.~\cite{WANG202111}, the improved fully-connected neural network architecture proposed in Ref.~\cite{gradientpathologies}  and the integrated neural network architecture developed in this paper.

%\wy{On the other hand, for some special multi-scale problems, such as high-frequency and multi-frequency function, spectral bias impedes the accurate approximation of the multi-scale function for conventional fully-connected neural network \cite{WANG200022110768}. To this end, Wang et al.~\cite{WANG202111} construct novel architectures by Fourier feature embedding.} 
%\wy{But, a well-designed architecture alone does not guarantee accurate predictions.}
%\wy{So, it is needed to combine the existing state-of-the-art architecture with the reasonable regularization methods to enhance prediction accuracy when solving the special multi-scale problems, which is another motivation of this paper.}

% \wy{
% \begin{remark}
%     It is noted that to avoid the difficulty caused by the large difference between the supervised term and the loss term, the hard constraint approach \cite{w712178,Moseley}, which strictly enforce the boundary conditions by using the neural network as
% part of a solution ansatz, looks like a good way. However, our numerical experiments show that such an approach does not work effectively for multi-scale problems.
% \end{remark}
% }

\subsection{A regularization strategy for multi-magnitude loss terms}\label{normstra}

The word ``regularization'' here means processing the values of the loss terms to approximately the same order of magnitude.
In order to make each loss term of the loss function to be optimized synchronously in the optimization process, one can use the corresponding mechanism provided by the conventional PINN to reduce the difference in magnitude between different loss terms by adjusting their weights ~\cite{METHODADUALDIMERMETHOD,Self-adaptivepinnjcp,dasapinnpinnpinn}. The weight selection is related to specific problems and has a wide range of values, so it is very difficult to give an exact selection criteria.

Consider the following formula:
\begin{equation}\label{nroot}
   \lim_{n\to\infty}{a^{\frac{1}{n}}}  = 1, \quad \forall a\ge \epsilon > 0.
\end{equation}
If $a=10^9, b=10^3$, the difference between them is 6 orders of magnitude. But $a^\frac{1}{3}$ has the same magnitude as $b$.
Inspired by this, we can reduce the difference in magnitude between the loss terms by computing their $\frac{1}{n}$ power, for which we define the following loss function:
\begin{equation}\label{newLF}
\mathcal{\Tilde{L}}(\theta;\Sigma)=w_s \mathcal{L}^{\frac{1}{m}}_s(\theta;\tau_{s})+w_r \mathcal{L}^{\frac{1}{n}}_r(\theta;\tau_{r}), \quad m>0, n>0,
\end{equation}
where $m$ and $n$ are called regularization parameters.
Clearly, if $\mathcal{\Tilde{L}}(\theta;\Sigma)$ tends to 0, then the standard loss function $\mathcal{L}(\theta;\Sigma)$ in Eq.~\eqref{loss_std} must also tend to 0.

We call the method of obtaining the prediction function for multi-scale problems with \textbf{multi-magnitude} loss terms by optimizing Eq.~\eqref{newLF} as the \textbf{MMPINN} method.
The new loss function has the following characteristics:
\begin{itemize}
    \item It is a generalization of the conventional PINN loss function, and 
    it degenerates to the standard loss function when $m=1$, $n=1$.

    \item By simply adjusting $m$ or $n$, one can regularize $\mathcal{L}_s(\theta;\tau_{s})$ and $\mathcal{L}_r(\theta;\tau_{r})$ to approximately the same order of magnitude, which allows the optimization algorithm to optimize both simultaneously rather than favoring one over the other.

    \item According to Eq.~\eqref{nroot}, the choice of $m$ and $n$ can be independent of the problem, and it is always possible to make $\mathcal{L}_s(\theta;\tau_{s})$ and $\mathcal{L}_r(\theta;\tau_{r})$  approximately the same magnitude as long as $m$ and $n$ are large enough.
    \item Depending on the actual order of magnitude of $\mathcal{L}_s(\theta;\tau_{s})$ and $\mathcal{L}_r(\theta;\tau_{r})$, one can suppress or accelerate the optimization of the corresponding loss terms to achieve the desired effect by adjusting $m$ or $n$ during the training process.

    % \item By including the balance parameters M and N, the distribution of internal residual losses can be better balanced, allowing higher accuracy to be achieved.
    
    % \item For sub-scale problems, M and N are adjusted to find the optimal parameters of the neural network $u(x,t;\theta)$ in a more rational way while training.
\end{itemize}

\begin{remark}
To reduce the order of magnitude of the loss terms, another natural idea is to use a logarithmic function for the loss terms, which is a common approach in the field of data processing. This approach can reduce the order of magnitude of the loss terms to 1 and obviates the need to tune  hyper-parameters such as $m$ or $n$. However, the logarithmic approach has two problems: 
(1) The value range of the logarithmic function is $[-\infty, +\infty]$, which is inconsistent with the value range of the original loss function; 
(2) Even if we consider using the logarithmic function only for the loss terms greater than 1, because the logarithmic function excessively compresses the value of the loss function,  the gradient of the corresponding loss term becomes extremely small, resulting in a case similar to the disappearance of the gradient, which causes the optimizer to obtain an invalid prediction result.
\end{remark}

\subsection{Multi-level training algorithms}\label{Multi-level training strategy for multi-scale models}
Theoretically, if $\Tilde{\mathcal{L}}(\theta;\Sigma) \to 0$, we can get $\mathcal{L}(\theta;\Sigma) \to 0$. 
Nevertheless, when the two loss functions are optimized by an optimization algorithm, the optimization process and the optimization results may be significantly different due to the different properties of the two functions.

When using Eq.~\eqref{newLF} to handle loss terms of varying magnitudes,  the small magnitude loss terms could be synchronously optimized. However, the large magnitude loss terms may not be fully optimized due to suppression. To address this problem, we propose the \emph{multi-level training algorithm}, which ensures that the optimized loss function of the new algorithm exactly matches with that of the conventional PINN method.
This theoretically ensures that the results of the two methods are completely equivalent.

To clearly describe the multi-level training strategy, we assume that $\mathcal{L}_r(\theta;\tau_{r})$ is a loss term with large orders of magnitude and let $m=1, n=3$. For this case, we can obtain different levels of prediction by the following steps:
\begin{itemize}
    \item \textit{Level 1}
    
Take the random parameters  $\theta_0$ as the initial network parameters and use the combined Adam and L-BFGS optimizers to optimize the following loss function
\begin{equation}\label{levelOne}
\mathcal{\Tilde{L}}_1(\theta;\Sigma)=w_s \mathcal{L}_s(\theta;\tau_{s})+w_r \mathcal{L}^{\frac{1}{3}}_r(\theta;\tau_{r}),
\end{equation} 
then we get $\theta_1$ and the first level of the prediction function $u_{\theta_1}$. 

    \item \textit{Level 2}

Utilize the $\theta_1$ as the initial guess of the neural network parameter
and use the L-BFGS optimizer to optimize the following loss function
\begin{equation}\label{leveltwo}
\mathcal{\Tilde{L}}_2(\theta;{\theta_1},\Sigma)=w_s \mathcal{L}_s(\theta;{\theta_1},\tau_{s})+w_r \mathcal{L}^{\frac{1}{2}}_r(\theta;{\theta_1},\tau_{r}),
\end{equation} 
then we get $\theta_2$ and the second level of the prediction function $u_{\theta_2}$. 

    \item \textit{Level 3}

Utilize $\theta_2$ to initialize network parameter and use the L-BFGS optimizer to optimize the following loss function
\begin{equation}\label{levelthird}
\mathcal{\Tilde{L}}_3(\theta;{\theta_2},\Sigma)=w_s \mathcal{L}_s(\theta;{\theta_2},\tau_{s})+w_r \mathcal{L}_r(\theta;{\theta_2},\tau_{r}),
\end{equation} 
then we get $\theta_3$ and the third level of the prediction function $u_{\theta_3}$, which is also an approximation to the solution of Eq.~\eqref{pde function}. 
    \end{itemize}

The above multi-level algorithm is only designed to theoretically achieve the same results as the conventional PINN, and practical experience shows that in most cases the results of the first level prediction are sufficient. In addition, the algorithm can also jump directly from the first level to the third level without necessarily going through the second level.

For the multi-level algorithm, MMPINN can be considered as a pre-training strategy \cite{GUO2023112258} of the conventional PINN method to obtain a set of good initial guess of neural network parameters for the optimizer. In most cases, this pre-training strategy is so effective that a few iterations are required to optimize the loss function of the conventional PINN using it as the initial guess of the network parameters.

\subsection{Grouping regularization approach for different subdomains}\label{groupregstra}
Generally, the solution of the governing equation exhibits varying features across different locations in the computational domain.
Significant variations in the solution's gradient across different locations of the computational domain may lead to widely varying residual magnitudes on different subdomains.
In this case, combining them together for optimization according to Eq.~\eqref{Lr_loss} necessarily results in the subdomains with smaller residual magnitude not being optimized synchronously. To solve this problem, we present the \emph{grouping regularization approach}.

\begin{figure}[h]
    \centering
\includegraphics[width=0.5\textwidth]{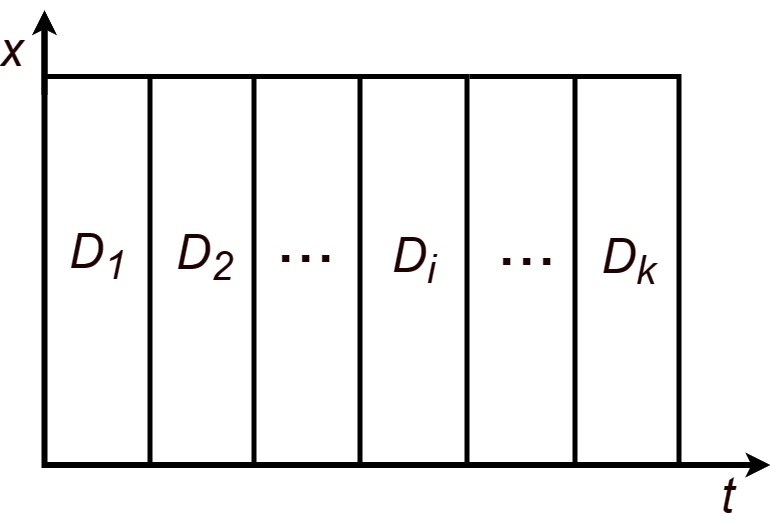}
    \caption{{Grouping along the time direction}.}
    \label{fig:GroupingwayP1}
\end{figure}

Taking the evolution equation as an example, assuming that the property of the solution of the equation has a large difference in the time direction, we divide the whole time domain $[0,T]$ into $k$ time-subdomains, i.e,
\begin{align*}
    [0,T]
=[0,T_1]\cup(T_1,T_2]\cup\dots\cup(T_{k-1}, T_k].
\end{align*}
Figure \ref{fig:GroupingwayP1} is a schematic diagram of the division of the entire domain into several subdomains along the temporal axis.

Based on this grouping way, the loss function is constructed as follows:
\begin{equation}\label{grouploss}
\begin{cases}
\mathcal{L}(\theta;\Sigma)=w_s\mathcal{L}_s(\theta;\tau_{s})+\mathcal{L}_r(\theta;\tau_{r}),\\
\mathcal{L}_{s}(\theta;\tau_{s}) = \frac{1}{N_s}\sum_{i=1}^{N_s}\left|\mathcal{B}(u_\theta(x_i))-g(x_i)\right|^2,\\
\mathcal{L}_{r}(\theta;\tau_{r}) =\sum_{h=1}^{k}\omega_{r,h}{L}_{r,h}(\theta;\tau_{r,h}),\\

{L}_{r,h}(\theta;\tau_{r,h}) =\frac{1}{N_{r,h}} \sum_{i=1}^{N_{r,h}}\left| \mathcal{P}(u_\theta(x_{h,i}))-f(x_{h,i})   \right|^2,\
\end{cases}
\end{equation}
where $x_{h,i}=(x_i, t_{h,i})$ represents a sample point in the subdomain $D_h$, $\tau_{r,h}$ is the set consisting of these points, and
${L}_{r,h}(\theta;\tau_{r,h})$ is the residual term defined on the subdomain $D_h$.

In the same way as in Section \ref{normstra}, we set a regularization parameter for each subdomain, e.g., assign $n_h$ to the subdomain $D_h$:
\begin{equation}\label{grouploss-function}
\mathcal{L}_{r}(\theta;\tau_{r}) =\sum_{h=1}^{k}\omega_{r,h}[L_{r,h}(\theta;\tau_{r,h})]^{\frac{1}{{ n}_h}}.
\end{equation}
By reasonably setting the parameters $n_h$, we can focus more on some specific subdomain to get better prediction results when training the neural network.

Note that, the above is a grouping strategy for the whole computational domain along the temporal axis, of course it is also possible to divide the whole domain into several subdomains by spatial locations. The number of subdomains and the way they are grouped depends on the specific problem. 
%The basic guideline is: Domains whose residuals are of the same order of magnitude are grouped together.

% \subsection{Basic procedure of MMPINN method}
% The algorithm \ref{AL1} shows the basic implementation steps of the MMPINN framework, which is helpful for programming.
% \begin{algorithm}
% \caption{MMPINN Framework} \label{AL1}
% \begin{large}
% \BlankLine
%    \KwIn{\\
%         \qquad Neural network structure; \\
%         \qquad Balance parameters $M$,$N$;\\} 
% \KwOut{\\
%         \qquad Neural network parameters $\theta$\\}
% Define $m=M,n=N$;\\
% Neural network parameters initialization;\\
% Construction of the loss function: ~$\mathcal{L}(\theta;\Sigma)=w_s\mathcal{L}_s^{\frac{1}{m}
% }(\theta;\tau_{s})+w_r\mathcal{L}_r^{\frac{1}{n}}(\theta;\tau_{r})$;\\
% Optimisation of the loss function $\mathcal{L}(\theta;\Sigma)$ by Adam;\\
% Optimisation of the loss function $\mathcal{L}(\theta;\Sigma)$ by L-BFGS;\\
% \While{true}{
% \If{$m=1\:\&\&\:n=1$}{
% break;\\}
% \If{$m>1\:\&\&\:n=1$}{
% $m=m-1$;\\}
% \If{$m=1\:\&\&\:n>1$}{
% $n=n-1$;\\}
% \If{$m>1\: \&\& \:n>1$}{
% $m=m-1,n=n-1$;\\}
% Get the last neural network parameters;\\
% Construction of the loss function: ~$\mathcal{L}(\theta;\Sigma)=w_s\mathcal{L}_s^{\frac{1}{m}}(\theta;\tau_{s})+w_r\mathcal{L}_r^{\frac{1}{n}}(\theta;\tau_{r})$;\\
% Optimisation of the loss function $\mathcal{L}(\theta;\Sigma)$ by L-BFGS;\\

% }
% \end{large}
% \end{algorithm}

\subsection{An integrated neural network architecture}\label{subsec:Integrated neural network architectures}

Network architecture is important for the successful implementation of PINN methods in scientific computing \cite{gradientpathologies}.
As discussed in Section \ref{pro22}, the conventional deep neural networks have relatively weak expressiveness for multi-frequency problems due to spectral bias.
Therefore, it is necessary to study special network architecture to handle the typical multi-scale problem with multi-frequency, especially high-frequency features.

Through in-depth investigation, we find that the Fourier feature neural networks in Ref.~\cite{WANG202111} and the improved fully-connected neural networks in Ref.~\cite{gradientpathologies}
have their own merits in dealing with multi-scale problems. However, for multi-frequency problems with multi-scale loss terms, using either alone does not yield ideal results.
To solve arbitrary types of multi-scale problems, we develop an integrated network architecture by combining these two network architectures, which takes advantage of the characteristics of both and can deal relatively well with difficult multi-scale problems.

\begin{figure}[h]
    \centering
    \includegraphics[width=0.9\textwidth]{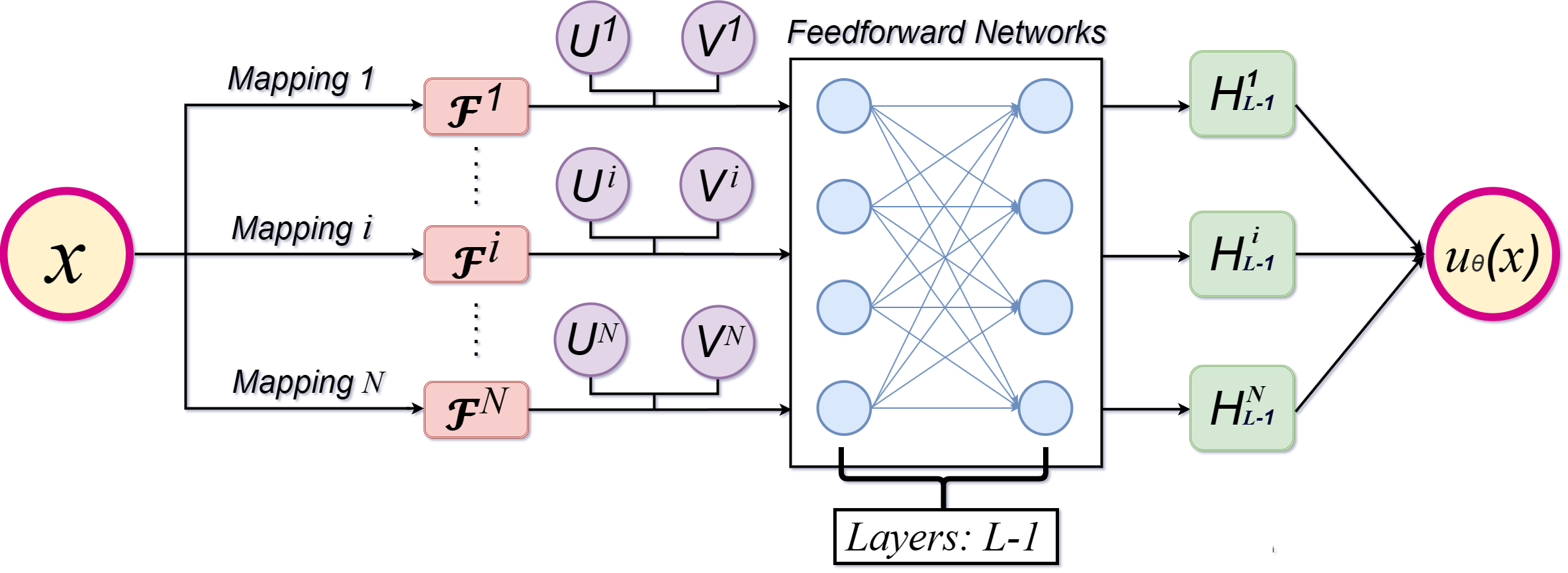} 
    \caption{INN network architecture.}
    \label{fig:INN-arch}
\end{figure}

The design idea of the integrated neural network architecture is as follows:
First, we apply multiple Fourier feature embeddings to the input coordinates. Second, we construct two transform networks for each Fourier feature. 
Third, we pass transform networks to the fully-connected neural networks. Finally, we concatenate the outputs with a linear layer.
Figure~\ref{fig:INN-arch} shows the schematic of the integrated neural network architecture, which we will formulate in detail next.

%The detailed integrated neural network architecture is as follows. 
The Fourier mapping $\mathcal{F}$ is defined as
\begin{equation}\label{Fourier mapping}
\mathcal{F}^{i}(x) = \left [ \begin{matrix}
\sin (B^{(i)}x) \\
\cos (B^{(i)}x) \\
\end{matrix} \right ],
%\quad H_{x,1}^{(i)}= \phi(W_1 \cdot f^{(i)}_x(x) +b_1 ),
\quad i=1,2,\cdots,N,\\
\end{equation}
where $N$ represents the number of Fourier mapping for the input coordinates $x$. Each entry of $B^{(i)}$  is sampled from the Gaussian distribution $\mathcal{N}(0,(\sigma_i)^2)$ and is held fixed during model training.
When $\sigma=1$, low frequencies are learned initially, which is similar to the behavior of deep neural networks.  If $\sigma$ is larger, high frequencies are learned first, which may result in over-fitting \cite{WANG202111}. Therefore, selecting the appropriate $\sigma$ is crucial.
The effectiveness of the Fourier feature embeddings in mitigating the spectral bias is illustrated in Ref.~\cite{WANG202111}.
Using this technique can improve the accuracy of solving multi-frequency problems.

The two transform networks are defined as
\begin{equation}\label{two-transform-network}
\begin{cases}
U^i=\phi(W_u^{i} \cdot \mathcal{F}^{i}(x) + b_{u}^{i}),\\
V^i=\phi(W_v^{i} \cdot \mathcal{F}^{i}(x) + b_{v}^{i}),& i=1,2,\cdots,N,\\
\end{cases}
\end{equation}
where $\phi$ denotes a nonlinear activation function and $(W_{u}^{i}, b_{u}^{i}, W_{v}^{i}, b_{v}^{i})$ are the parameters to be optimized.
The transform networks based on the Fourier mapping, formulated by Eq.~(\ref{two-transform-network}), are different from those in Ref.~\cite{gradientpathologies}. They are the improved version by replacing the input variable $x$ with $\mathcal{F}^{i}(x)$.
Introducing the two networks into conventional fully-connected neural network architecture could enhance the hidden states with residual connections to alleviate the stiffness of the the gradient flow \cite{gradientpathologies}. 

The subsequent  detailed forward pass is
\begin{equation}\label{subsequent  detailed forward pass}
\begin{cases}
Z_{1}^{i}=\phi(W_1 \cdot \mathcal{F}^{i}(x) +b_1 ),\\
H_{1}^{i}=(1- Z_{1}^{i} )\odot U^i+Z_{1}^{i}\odot V^i,\\
Z_{l}^{i}=\phi(W_l \cdot H_{l-1}^{i} +b_l ),\\
H_{l}^{i}=(1- Z_{l}^{i} )\odot U^i+Z_{l}^{i}\odot V^i,
&i=1,2,\cdots,N,\quad l=2,\cdots,L-1,\\
\end{cases}
\end{equation}
where $\odot$ represents the point-wise multiplication and $L$ is the number of hidden layers of the neural network. 
The final output is
\begin{equation}\label{The final output}
u_\theta(x)=W_{L}\cdot[H_{L-1}^{1},\cdots,H_{L-1}^{N}]+b_{L}.\\
\end{equation}
The weights $W_l$ and the biases $b_l$ in Eq.~\eqref{subsequent  detailed forward pass} and Eq.~\eqref{The final output} are also the parameters of the neural network to be obtained by optimizing the loss function.
%We will validate the effectiveness of the proposed architecture through a typical systematic numerical experiment in Sect.~\ref{subsec:kgsection}.

In Ref.~\cite{WANG202111}, the authors also propose a \textbf{spatio-temporal} multi-scale Fourier mapping, which is similar to the Fourier mapping~(\ref{Fourier mapping}) and can also be combined with the improved fully-connected neural networks in Ref.~\cite{gradientpathologies} to form a new network architecture. 
For simplicity of description, we refer to the (spatio-temporal) Multi-scale Fourier Feature neural network architectures in Ref.~\cite{WANG202111} as MFF and the corresponding Integrated Neural Network architectures as INN in this paper.

By combining MMPINN with different network architectures, such as the commonly used deep neural networks, MFF and INN, we obtain several new PINN methods. 
Table \ref{Architecture of Neural Networks in MMPINN} lists the names of the new methods, and their performance is examined in Section \ref{sec:Numerical Examples}.

\begin{table}[h]
		\setlength{\abovecaptionskip}{0cm}
		\setlength{\belowcaptionskip}{0.2cm}
\caption{MMPINN with different network architectures.}
\label{Architecture of Neural Networks in MMPINN}
\centering
\begin{tabular}{c|c} 
\hline 
\textbf{Neural Network Architecture}& \textbf{Method}\\
\hline
common Deep Neural Network & MMPINN-DNN \\
\hline
Multi-scale Fourier Feature Network & MMPINN-MFF \\
\hline
Integrated Neural Network & MMPINN-INN \\
\hline
\end{tabular}
\end{table}

    Numerical experiments show that MMPINN-INN obtains highly accurate  computational results for difficult multi-scale problems with high-frequency features, which has obvious advantages over MMPINN-DNN and MMPINN-MFF. However, INN requires large computational resources, and improving its computational efficiency is one of our next works.

\subsection{MMPINN framework}\label{subsec: Multi-scale Fourier feature architecture}

\begin{figure}[h]
     \centering \includegraphics[width=1.0\textwidth]{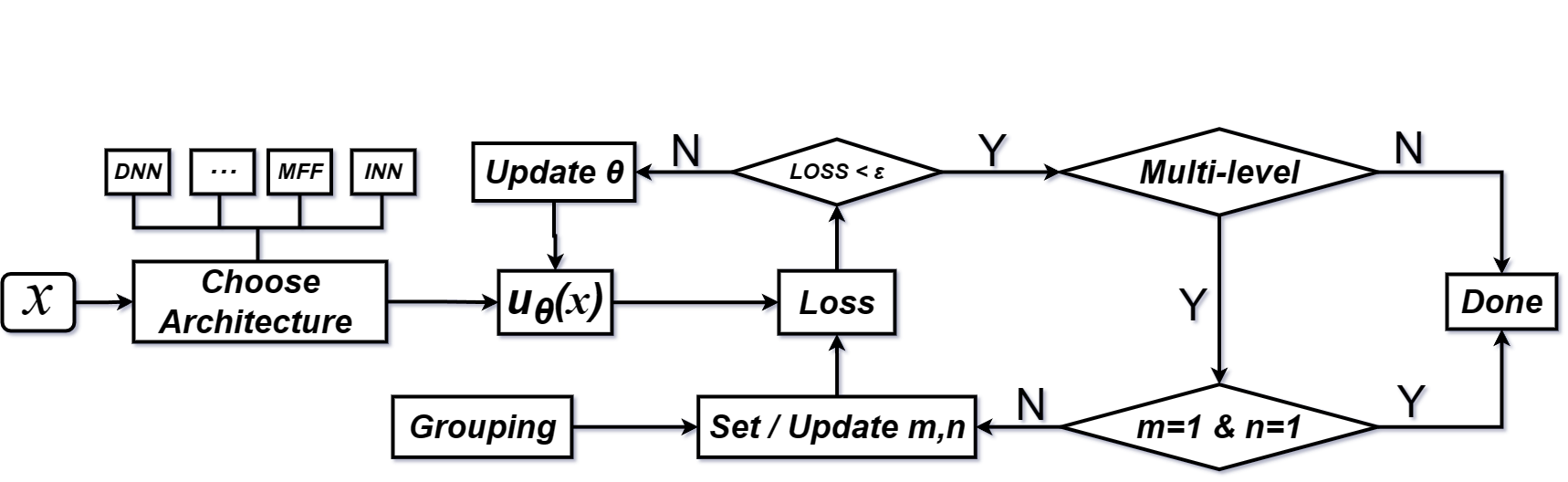} 
     \caption{Diagram of the MMPINN framework.}
    \label{fig:The diagram of the MMPINN framework}
 \end{figure}

By combining all of the above strategies, we construct an improved PINN framework called the MMPINN framework. Using this framework, it is possible to design deep learning methods that obtain highly accurate predictions for a large class of multi-scale problems with multi-magnitude loss terms and multi-frequency features.
Figure \ref{fig:The diagram of the MMPINN framework} is the diagram of the MMPINN framework.

% To well handle the multi-scale problems with multi-magnitude loss terms,
% \wy{we set the MMPINN framework using the conventional PINN framework in Section \ref{sec:std-pinn}, regularization strategy in Sect.~\ref{normstra}, multi-level training algorithms in Sect.~\ref{Multi-level training strategy for multi-scale models}, grouping regularization approach in Sect.~\ref{groupregstra}, and various network architectures in Sect.~\ref{subsec:Integrated neural network architectures}. Figure \ref{fig:The diagram of the MMPINN framework} illustrates the diagram of the MMPINN framework in detail, which will be further expanded in our future work.}

%Approximating a high-frequency or multi-frequency function with a conventional fully-connected neural network (CFNN) is not difficult if sufficient data is available within the computational domain. However, spectral bias \cite{pmlr-v97-rahaman19a} impedes the accurate approximation of the multi-scale function by conventional PINN methods \cite{WANG200022110768}. So, to mitigate the pathology, Wang et al.~\cite{WANG202111} propose spatio-temporal multi-scale Fourier feature (MFF) architecture by Fourier feature embedding. 

\section{Numerical Examples}\label{sec:Numerical Examples}

In this section, several numerical experiments are performed to validate the performance of our method. 
In Section~\ref{subsec:S41}, we show the robustness of the MMPINN-DNN method by solving a large gradient heat conduction problem.
The examples in Section~\ref{subsec:multi-frequency}, Section~\ref{subsec:high-frequency} and Section~\ref{subsec:kgsection} are taken from the recent literature, and we compare our method with the state-of-the-art methods. 
In Section~\ref{subsec:multi-scale}, we further investigate the performance of our method on a difficult multi-frequency model.
In Section~\ref{subsec:S44}, the grouping regularization strategy is used to solve a problem with dramatic variations in different subdomains, demonstrating its effectiveness.

%In Sect.~\ref{subsec:kgsection}, we make a comprehensive comparison between MMPINN-INN and  other methods mentioned in Ref.~\cite{gradientpathologies} for solving an nonlinear and time-dependent problem. 

%In Sect.~\ref{subsec:S43}, we test the performance of the MMPINN-DNN method on a strongly anisotropic diffusion equation.

The $L_2$ relative error norm is used to measure the accuracy of the prediction, which is defined as follows:
\begin{equation}\label{L2norm}
\left\|\epsilon\right\|_2=\frac{\sqrt{\sum_{i=1}^N\left|u_\theta(x_i)-u(x_i)\right|^2}}{\sqrt{\sum_{i=1}^N\left|u(x_i)\right|^2}},
\end{equation}
where $u(x_i)$ is the real solution or the reference solution, and $u_\theta(x_i)$ is the neural network prediction for a point in the test set $\{x_i \mid _{i=1}^{N} \in \Omega \}$. $N$ is the size of test set.

We use the deep learning framework TensorFlow (version 1.13.1 or 2.11.0)  to implement all experiments.
The data type is \texttt{float32} and the activation function is \texttt{tanh} for all examples. 
We combine Adam and L-BFGS to optimize the loss function, without any special statement, first running 2000 Adam iterations and then switching to L-BFGS iterations until convergence. 
For our method, 2000 Adam iterations are sufficient to obtain available predictions; however, for the conventional PINN method, this setting often fails to produce available predictions, and thus it is necessary to run $10^6$ Adam iterations and then switch to L-BFGS until convergence, which will be referred to as conventional PINN ($10^6$ Adam) in the following sections.
All parameters and termination criteria of the L-BFGS optimizer are considered as suggested in Refs.~\cite{Self-adaptivepinnjcp,LLLBFGS}.
Before training, the parameters of the neural network are randomly initialized by using the Xavier scheme \cite{XuC}.
%The weights of the loss terms are generally set to $w_s=1$ and $w_r=1$.

% \begin{remark}\label{remark4.1}
% It is worth noting that in our numerical experiments we only try to demonstrate the validity of the MMPINN framework, and we do not try to find the best computational accuracy by adjusting some hyperparameters.
% \end{remark}

\subsection{A heat conduction problem with large gradients}\label{subsec:S41}
In practical applications such as inertial confinement fusion~\cite{MRE22}, it is often necessary to solve the heat conduction problem with a strong heat source that suddenly appears at a certain time, and this example is designed to test the effectiveness of our method for this type of problem.

Consider the following heat equation
\begin{equation}\label{hcp1}
\begin{cases}
u_t=u_{xx}+f(x,t),& x\in(-1,1),t\in(0,1],\\
u(x,0)=(1-x^2)e^\frac{1}{1+\varepsilon},& x\in(-1,1),\\
u(\pm 1,t)=0,&t\in(0,1].\\
\end{cases}
\end{equation}
Its exact solution is given by
\begin{equation}\label{Exact_Ex1}
u(x,t)=(1-x^2)e^\frac{1}{(2t-1)^2+\varepsilon},
\end{equation}
where $\varepsilon$ is a positive constant, and $f(x,t)$ is derived from the exact solution.  

% Figure~\ref{fig:Exact solution n different values of value} shows the exact solution for three cases ($\varepsilon=0.3,0.15,0.11$), and we show the multiscale properties in different cases by the order of the initial losses $\mathcal{L}_r:\mathcal{L}_s$. We can see from the figure that, as the value of $\varepsilon$ decreases, the weak singularity of the equation around $x=\pm1$ becomes more pronounced and the gradients of the solution $u$ and the source term $f$  become larger, leading to an increasingly multiscale property of the equation. 

 % \begin{figure}[h]
 %     \centering
 %     \includegraphics[width=1\textwidth]{./fig/exact solutions to the heat conduction equation.pdf} 
 %     \caption{Exact solutions of Sect.~\ref{subsec:S41} for $\varepsilon=0.3,0.15,0.11$, and the initial losses $\mathcal{L}_r:\mathcal{L}_s=10^3:1$, $10^6:1$, and $10^8:1$, respectively. }
 %     \label{fig:Exact solution n different values of value}
 % \end{figure}

For this model, if $\epsilon$ is large, the variation of $u$ will be smooth over the entire computational domain; if $\epsilon$ is very small, the value of $u$ will change sharply near $t=0.5$, embodying typical multi-scale characteristics. 
Since the boundary value of this model is always 0 and its initial value is less than $e$, the value of its supervised loss term is always small, while the value of its residual term increases significantly as $\epsilon$ decreases.
For different $\varepsilon$, the ratio of initial losses of the two terms from practical computation is as follows:
\begin{align*}
    &\varepsilon=0.3,\enspace \quad \mathcal{L}_s:\mathcal{L}_r=1:10^3,\\
    &\varepsilon=0.15, \quad \mathcal{L}_s:\mathcal{L}_r=1:10^6.
    % &\varepsilon=0.11, & \mathcal{L}_s:\mathcal{L}_r=1:10^8.\\
\end{align*}
As we know, the conventional PINN method can usually achieve good predictions for smooth problems, while for the problem with large difference in magnitude between the supervised term and the residual term, it is difficult to obtain valid predictions without fine-tuning hyperparameters.

%To reduce the significant difference in magnitude between the supervised term and the residual term, one option is to assign their respective weights as the inverse of the initial loss \cite{xie2022weighted}.The weights of the loss terms are set to $w_s=1$ and $w_r=1e-6$, which is referred to as \wy{weighted PINN method} in Table \ref{table2_ex1} for solving Eq.~\eqref{hcp1} with $\varepsilon=0.15$.
We choose the MMPINN-DNN method with $m=1$, $n=3$ to solve this model for different $\varepsilon$, and compare its result with those of conventional PINN methods and the SA-PINN method in Ref.~\cite{Self-adaptivepinnjcp}. Table \ref{table2_ex1} lists the $L_2$ relative errors of various methods.
%As shown in Table \ref{table2_ex1}, 
\emph{Using 2000 Adam iterations + L-BFGS},
the conventional PINN can give an accurate prediction for $\varepsilon=0.3$, 
but can no longer give a correct prediction for $\varepsilon=0.15$, while
the MMPINN-DNN method can still give an accurate prediction.
To obtain an available prediction with the conventional PINN method, we have to let it run $10^6$ Adam iterations.  

\begin{table}[h]
		\setlength{\abovecaptionskip}{0cm}
		\setlength{\belowcaptionskip}{0.2cm}
\caption{The $L_2$ relative errors of different methods for solving Eq.~\eqref{hcp1}.}
\label{table2_ex1}
\centering
\begin{tabular}{c|cc}
\hline
$\varepsilon$&Method&  $\left\|\epsilon\right\|_2$ \\
\hline
\multirow{2}{*}{\makecell[c]{$0.3$}} & 
conventional PINN&$3.41\pm 1.71\times 10^{-4}$ \\
&MMPINN-DNN (this work) &$1.86\pm 0.61\times 10^{-4}$ \\
\hline
\multirow{6}{*}{\makecell[c]{$0.15$}} & 
conventional PINN&$1.07\pm 0.11\times 10^{0}$ \\
&Hard Constraints~\cite{w712178} PINN &$1.10\pm 0.12\times 10^{0}$ \\
%&Weighted PINN &$4.47\pm 1.51\times 10^{-1}$ \\
&conventional PINN ($10^6$ Adam) &$5.26\pm 3.43\times 10^{-3}$ \\
&SA-PINN \cite{Self-adaptivepinnjcp} &$1.76\times 10^{-3}$ \\
&MMPINN-DNN (this work)& $5.01\pm 1.52\times 10^{-4}$ \\
\hline
\end{tabular}
\end{table}

\begin{figure}[!h]
    \centering
    \includegraphics[width=1\textwidth]{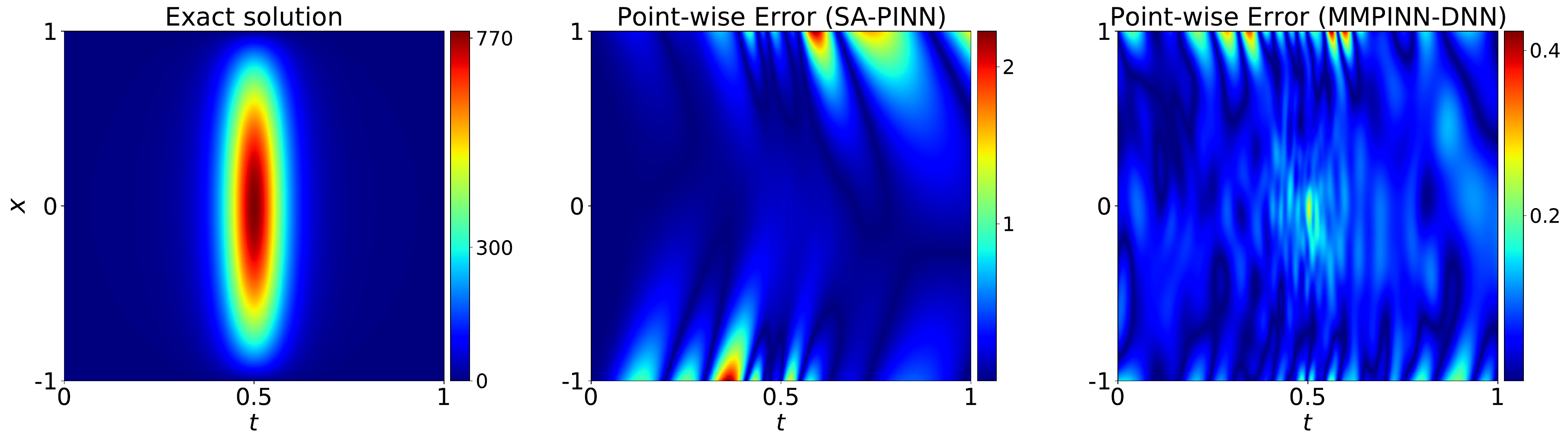} 
    \caption{The exact solution of Eq.~\eqref{hcp1} with $\varepsilon=0.15$ and the point-wise absolute errors of the SA-PINN \cite{Self-adaptivepinnjcp} and MMPINN-DNN methods.}
    \label{fig:0.15 exact solution and point-wise error by PINN and MMPINN}
\end{figure}

\begin{figure}[h]
     \centering \includegraphics[width=0.72\textwidth]{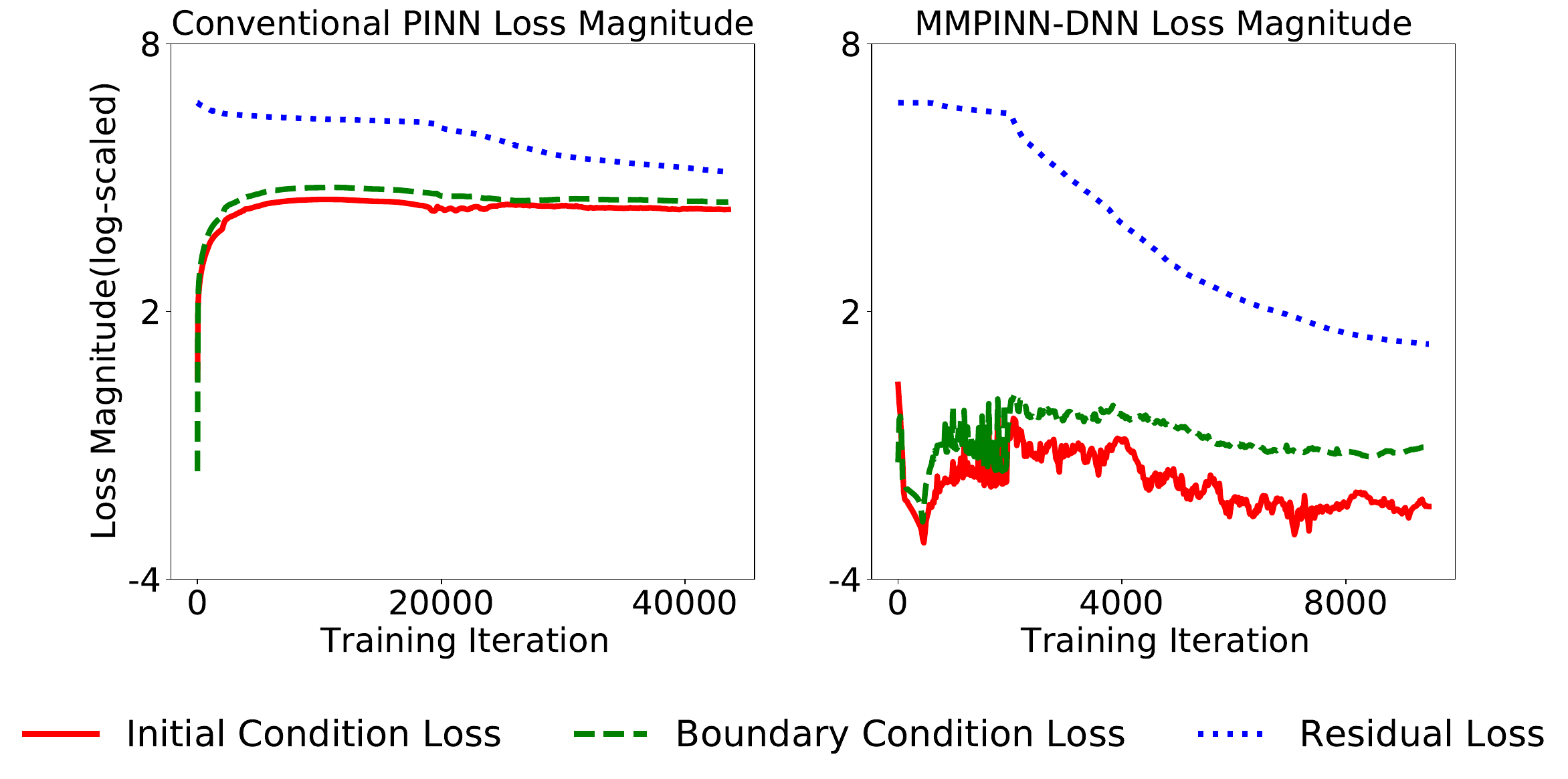} 
     \caption{Loss curves of the conventional PINN method and the MMPINN-DNN method for solving Eq.~\eqref{hcp1} with $\varepsilon=0.15$.}
     \label{fig:Loss_hist}
 \end{figure}

Figure \ref{fig:0.15 exact solution and point-wise error by PINN and MMPINN} shows the exact solution of Eq.~\eqref{hcp1} with $\varepsilon=0.15$ and the point-wise absolute errors of the SA-PINN \cite{Self-adaptivepinnjcp} and MMPINN-DNN methods.
%The MMPINN-DNN method is much better than conventional PINN methods not only in terms of computational accuracy, but also in terms of computational efficiency.
According to Table \ref{table2_ex1} and Figure \ref{fig:0.15 exact solution and point-wise error by PINN and MMPINN}, we can see that neither adjusting the weights \cite{Self-adaptivepinnjcp} nor increasing the number of optimizations could yield better results than the MMPINN-DNN method.
Figure \ref{fig:Loss_hist} shows the variation curves of different loss terms during the optimization process (2000 Adam iterations + L-BFGS). 
The boundary conditon loss and the initial condition loss of the conventional PINN method increase instead of decreasing during the optimization process, which ultimately leads to training failure; the MMPINN-DNN method can simultaneously reduce the boundary loss, initial condition loss and residual loss, which ensures accurate prediction results.

\begin{figure}[!h]
     \centering \includegraphics[width=0.56\textwidth]{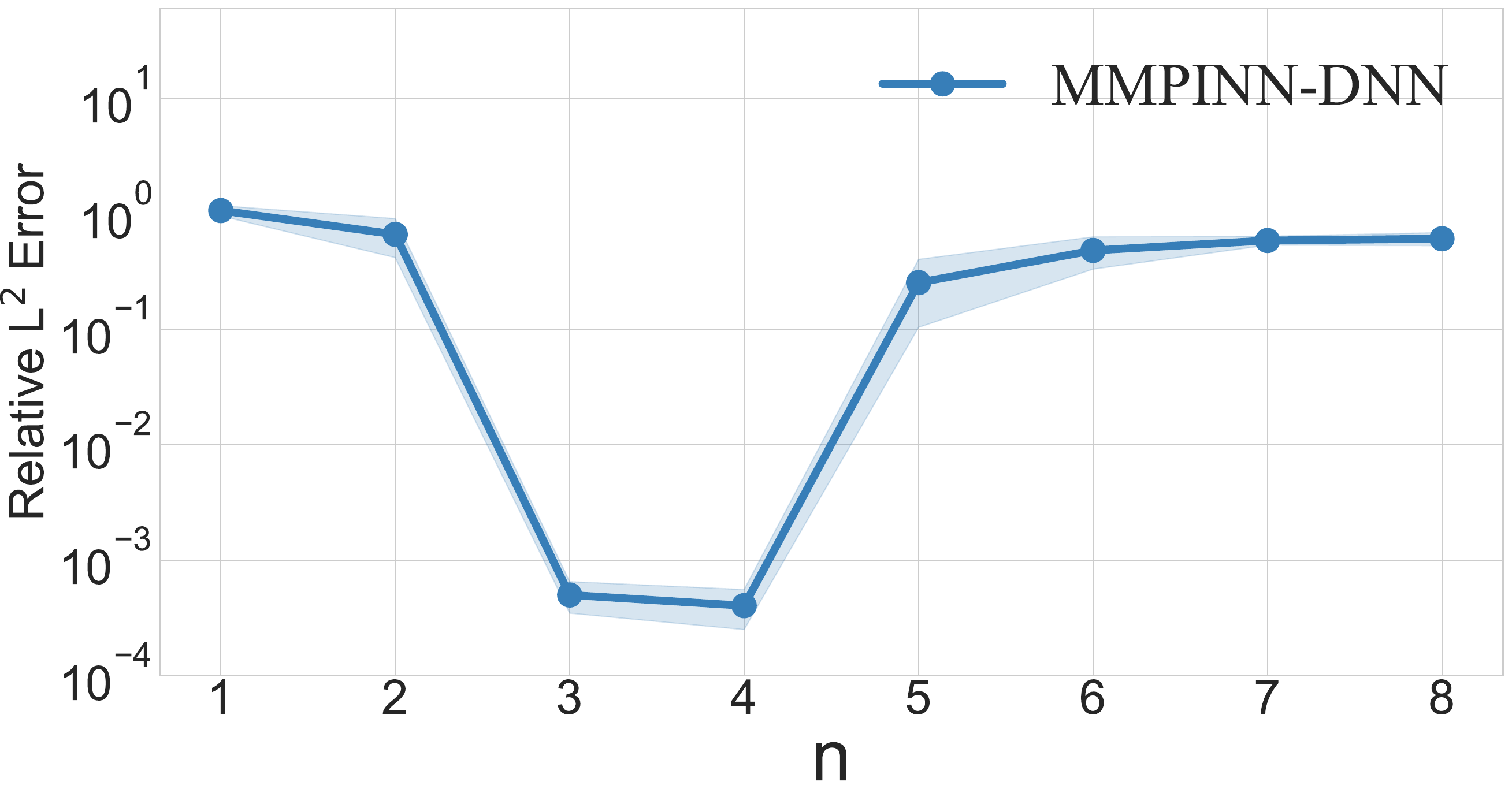} 
     \caption{The relative $L_2$ error using the MMPINN-DNN method with $m=1$ and $n=1,\cdots,8$ for solving Eq.~\eqref{hcp1} with $\varepsilon=0.15$. The line and the shaded region represent the mean and the standard deviation of 5 independent experiments, respectively.}
     \label{fig:different_hp}
 \end{figure}

To investigate the effect of the regularization parameters on the computation results, we test the cases of $n=1,\cdots,8$. Meanwhile, to avoid randomness, each set of tests is performed five times independently. The test results are shown in Figure~\ref{fig:different_hp}.
We see that the parameter $n$ cannot be chosen too large; large $n$ will excessively inhibit the residual term to be reasonably optimized, thus failing to obtain valid predictions satisfying the equation.
%It can be observed that choosing the proper parameters (m,n) is essential for the performance of the MMPINN-DNN method.If $n\le 2$, the problem that the significant difference in magnitude between the supervised term and the residual term could not be solved.Conversely, if $n \ge 5$,  the difference in magnitude between the loss terms  is entirely eliminated and the $L_2$ error is very large. The reason is probably that the small differences in magnitude between the loss terms are likely to contain relevant physical information about the equations, aiding the neural network in learning the correct solution. That is the reason that  we don't construct loss function using  logarithm  that frequently employed in the machine learning domain.}
 
\begin{figure}[!h]
    \centering
    \includegraphics[width=1\textwidth]{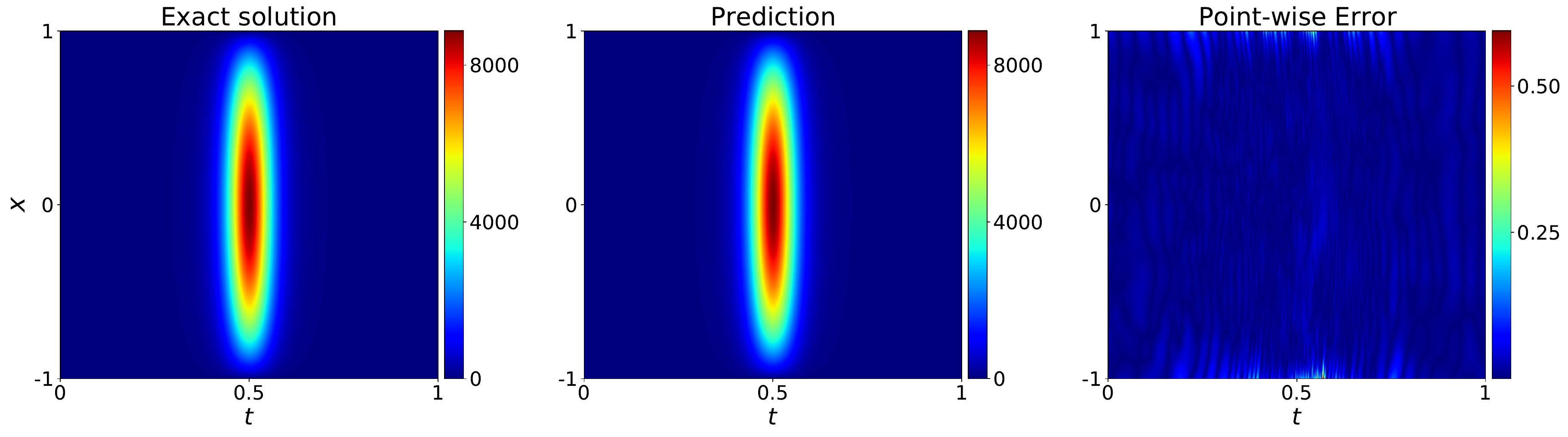} 
    \caption{The exact solution of Eq.~\eqref{hcp1} ($\varepsilon=0.11$) and the prediction and the point-wise absolute error of the MMPINN-DNN method.}
    \label{fig:0.11}
\end{figure}

We further test the performance of the MMPINN-DNN method in solving Eq.~\eqref{hcp1} with smaller $\varepsilon$, and the results of 5 independently repeated experiments are shown in Table \ref{table3_ex1}.
In the cases where the multi-scale property is more significant, the MMPINN-DNN method can still give accurate predictions, but the conventional PINN methods cannot give usable predictions.
Figure \ref{fig:0.11} shows the result of the MMPINN-DNN method for the case $\epsilon=0.11$, where the ratio of the supervised loss term to the residual loss term differs by 8 orders of magnitude, i.e.~$\mathcal{L}_s:\mathcal{L}_r=1:10^8$.
For this model, the MMPINN-DNN method still gives a very accurate prediction. In the case where the maximum value of $u$ is close to $9000$, the maximum prediction error is less than $0.7$.
\begin{table}[h]
		\setlength{\abovecaptionskip}{0cm}
		\setlength{\belowcaptionskip}{0.2cm}
\caption{$L_2$ errors of the MMPINN-DNN method for Eq.~\eqref{hcp1} with different $\varepsilon$.}
\label{table3_ex1}
\centering
\begin{tabular}{c|c} 
\hline 
$\varepsilon$&$\left\|\epsilon\right\|_2$  \\
\hline
0.14 & $1.43\pm 0.39\times 10^{-4}$ \\
0.13 & $3.03\pm 0.38\times 10^{-5}$ \\
0.12 & $2.34\pm 0.54\times 10^{-5}$\\
0.11 & $2.06\pm 1.28\times 10^{-5}$ \\
\hline
\end{tabular}
\end{table}

\begin{table}[!h]
		\setlength{\abovecaptionskip}{0cm}
		\setlength{\belowcaptionskip}{0.2cm}
\caption{Hyperparameters used for the different $\varepsilon$ in Sec.~\ref{subsec:S41}.}
\label{table1_ex1}
\centering
\begin{tabular}{c|ccccc} 
\hline 
$\varepsilon$&DNN Layers & DNN Neurons &$N_0$ &$N_b$ &$N_r$ \\
\hline
0.3 & 4 & 20 & 700 & 1400 &3000\\
0.15 & 4 & 50 & 1200 & 2400 &10000\\
0.14 & 4 & 100 & 1600 & 3200 &20000\\
0.13 & 4 & 320 & 2000 & 4000 &60000\\
0.12 & 4 & 400 & 3000 & 6000 &90000\\
0.11 & 4 & 400 & 4000 & 8000 &100000\\
\hline
\end{tabular}
\end{table}

As the complexity of the problem increases, we need to increase the number of training points and the size of the neural network. Table \ref{table1_ex1} shows the different hyperparameters, where $N_0, N_b$ and $N_r$ are the number of initial, boundary and residual points, respectively.

\subsection{A typical multi-scale problem }\label{subsec:multi-frequency}

The example in this subsection is taken from Ref.~\cite{WANG202111}.
This is a classical and pedagogical model. Although this model is very simple in form, its solution exhibits low frequency in the macro-scale and high frequency in the micro-scale, much like many practical scenarios~\cite{WANG202111}.  

Consider the 1D Poisson equation as follows:
\begin{equation}\label{multi-frequency}
\begin{cases}
\Delta u(x)  = f(x),& x \in (0,1), \\
u(0)=u(1)=0,\\
\end{cases}
\end{equation}
where 
\begin{equation}\label{A typical multi-scale problem source term}
f(x)=-16\pi^2 \sin (4 \pi x) -2250 \pi^2  \sin (150 \pi x).
\end{equation}
The exact solution for this model is
\begin{equation}\label{exact multi-frequency solution}
u(x)=\sin (4 \pi x) +0.1 \sin (150 \pi x).
\end{equation}

Since this is a frequency-dependent multi-scale problem, we use the MPPINN-MFF method for comparison with the method in \cite{WANG202111}.
According to the ratio of the loss terms $\mathcal{L}_s:\mathcal{L}_r=10^{1}:10^{8}$, the regularization parameters are set with $m=3$ and $n=3$. 
The architecture uses 3 hidden layers with 100 neurons per hidden layer.
The Fourier feature mappings are initialized with $\sigma_1=1$ and $\sigma_2=25$, respectively.
Other hyperparameters are chosen as in Ref.~\cite{WANG202111}, including the exponential decay of the learning rate, the choice of the optimizer, etc., except for the batch sizes ($N_b=N_r=512$).
%The multi-level training strategy does not be employed due to only using the Adam optimizer.
%the conventional PINN ($10^6$ Adam) is trained using the Adam optimizer for one million iterations of gradient descent.

\begin{table}[h]	\setlength{\abovecaptionskip}{0cm}
		\setlength{\belowcaptionskip}{0.2cm}
\caption{The $L_2$ relative errors of different methods for solving Eq.~\eqref{multi-frequency}.}
\label{multifrequencyfrequencylist}
\centering
\begin{tabular}{c|c} 
\hline 
Method&  $\left\|\epsilon\right\|_2$ \\
\hline
conventional PINN& $3.24\pm 2.13 \times 10^{0}$\\ 
\hline
conventional PINN ($10^6$ Adam)& $4.25\pm 1.92\times 10^{0}$\\ 
\hline
MFF \cite{WANG202111}& $1.41\pm 1.43 \times 10^{-2}$\\ 
\cline{1-2}
Hard Constraints \cite{w712178}+MFF& $5.35\pm 2.62 \times 10^{-3}$\\ 
\cline{1-2}
%MMPINN-INN (this work)&$2.10\pm 0.79\times 10^{-3}$\\ 
%\cline{1-2}
MMPINN-MFF (this work)&$6.94\pm 1.37\times 10^{-4}$\\ 
\cline{1-2}
\end{tabular}
\end{table}

\begin{figure}[ht]
    \centering    \includegraphics[width=0.85\textwidth]{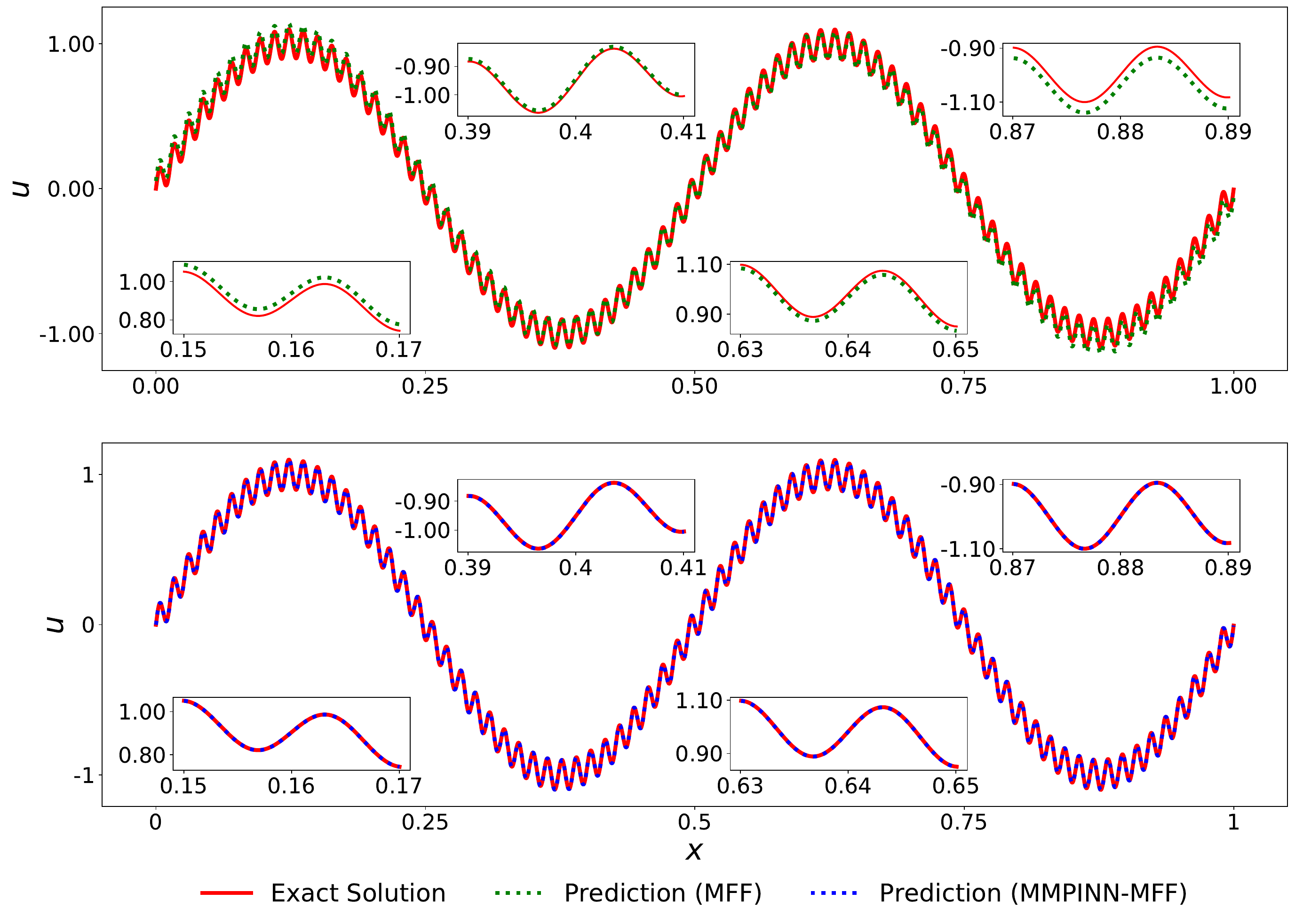}
    \caption{The exact solution and predictions for Eq.~\eqref{multi-frequency}. The top figure is the result of the MFF method and the bottom figure is the result of the MMPINN-MFF method.}
    \label{fig:taichangle}
\end{figure}

Table \ref{multifrequencyfrequencylist} lists the results of several PINN methods. From this table, we conclude that the deep neural networks cannot adequately learn the multi-frequency function even after $10^6$ iterations \cite{WANG202111}; Moreover, using the MFF alone does not guarantee strong robustness, its relative $L_2$ errors ranging from $0.41\%$ to $4.24\%$ in 5 independently repeated experiments. 
Our method simultaneously captures the low-frequency components and the high-frequency oscillations of the solution more accurately than the MFF method, as shown in Figure \ref{fig:taichangle}.

Note that the exact solution for Eq.~\eqref{multi-frequency} in Ref.~\cite{WANG202111} is given by
\begin{equation}\label{exact multi-frequency solution2}
u(x)=\sin (2 \pi x) +0.1 \sin (50 \pi x).
\end{equation}
Compared to Eq.~\eqref{exact multi-frequency solution}, the multi-scale feature of  Eq.~\eqref{exact multi-frequency solution2} is relatively small, and the the ratio of the initial loss terms is
$\mathcal{L}_s:\mathcal{L}_r=10^{1}:10^{6}$.
For this model, MFF is available and its relative $L_2$ error is $1.36\times 10^{-3}$ as reported in Ref.~\cite{WANG202111}, while the error of our method is $6.43\pm 2.52\times 10^{-4}$ under the same parameters.

\subsection{Helmholtz equation with high-frequency}\label{subsec:high-frequency}

The Helmholtz equation is an elliptic type partial differential equation describing electromagnetic waves. It has been used in many applications~\cite{lbpinnsXIANG202211}.
In Refs.~\cite{WANG202111,gradientpathologies,Self-adaptivepinnjcp}, the authors all take this equation as an example to test the performance of their approaches. 
The model they consider is the following:
\begin{equation}\label{high-frequency}
\begin{cases}
\Delta u(x,y) + k^2u(x,y) = q(x,y),& (x,y)\in \Omega = (-1,1)\times(-1,1), \\
u(x,y)=h(x,y),&(x,y)\in \partial \Omega.\\
\end{cases}
\end{equation}
The exact solution is assumed to be
\begin{equation}\label{exact high-frequency solution}
u(x,y)=\sin (a_1 \pi x)\sin (a_2 \pi y),
\end{equation}
$q(x,y)$ and $h(x,y)$ are derived from it.

%\begin{align}
%q(x,y) = & -(a_1 \pi)^2 \sin (a_1 \pi x) \sin (a_2 \pi y) \nonumber \\
%& -(a_2 \pi)^2 \sin (a_1 \pi x) \sin (a_2 \pi y) \nonumber \\
% + k^2 \sin (a_1 \pi x) \sin (a_2 \pi y).
%\end{align}

Making accurate predictions for this mode using the conventional PINN method is challenging \cite{gradientpathologies}. Due to a large initial difference in magnitude order between the residual term and the supervised term, the minimization of the residual term dominates the total training process.

Taking $k=1$, $a_1=1$ and $a_2=8$, we use this model to test our methods, including MMPINN-MFF and MMPINN-INN.
The network architecture contains 4 hidden layers, each layer with 50 neurons. Two separate Fourier feature mappings are initialized with $\sigma_1=1$ and $\sigma_2=10$, corresponding to $x$ and $y$, respectively. The numbers of training points are set with $N_0=1200$, $N_b=1200$ and $N_r=10000$.
The regularization parameters are set with $m=3$ and $n=3$, which are determined by the ratio $(\mathcal{L}_s:\mathcal{L}_r=10^{-1}:10^{5})$.

%On the other hand, selecting the appropriate neural network architecture is significant for accurately predicting high-frequency functions.

\begin{figure}[!h]
    \centering
    \includegraphics[width=1\textwidth]{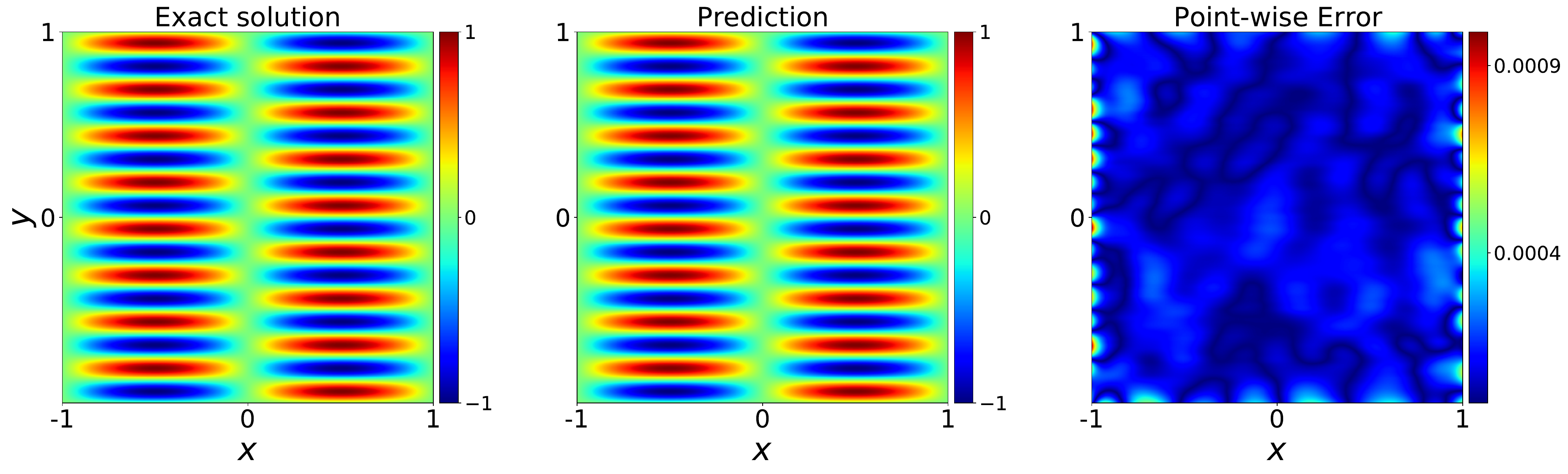} 
    \caption{The exact solution of Eq.~\eqref{high-frequency} and the prediction and the point-wise absolute error of the MMPINN-INN method.}
    \label{fig:Helmholtz_eqution_exact_solution_and_prediction}
\end{figure}

\begin{figure}[!h]
    \centering
    \includegraphics[width=1.0\textwidth]{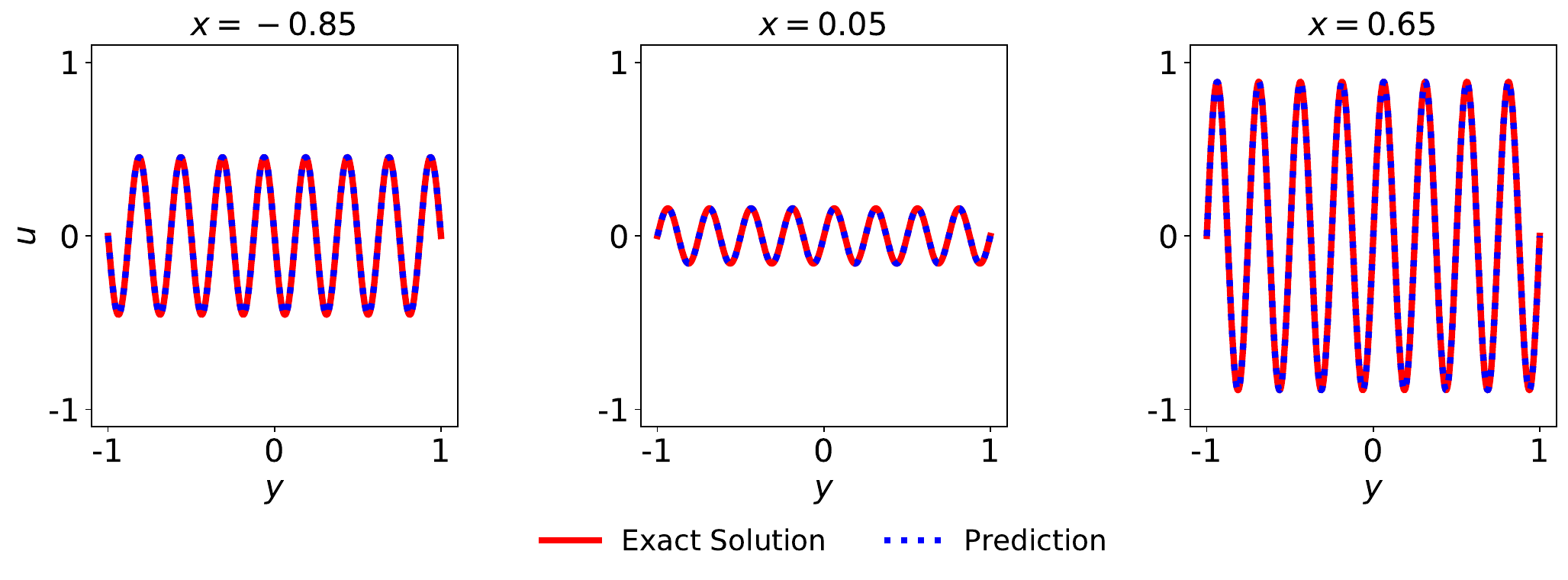} 
    \caption{Comparison of the exact solution and the prediction using MMPINN-INN.}
    \label{qswl}
\end{figure}

Table \ref{high-frequencylist} lists the results of different methods. Compared to the approaches in the recent literature, our approaches have better performance.
The MMPINN-INN method gives the best prediction among them, 
and its absolute error is shown in Figure \ref{fig:Helmholtz_eqution_exact_solution_and_prediction}.
Figure~\ref{qswl} shows the comparison of the exact solution and the prediction using the MMPINN-INN method at different positions.
The MMPINN-INN method captures the high frequency components of the solution very well.

\begin{table}[h]	\setlength{\abovecaptionskip}{0cm}
		\setlength{\belowcaptionskip}{0.2cm}
\caption{The $L_2$ relative errors of different methods for solving Eq.~\eqref{high-frequency}.}
\label{high-frequencylist}
\centering
\begin{tabular}{c|c} 
\hline 
Method&  $\left\|\epsilon\right\|_2$ \\
\cline{1-2}
LRA with IFNN~\cite{gradientpathologies}& $2.13\pm 0.68\times 10^{-2}$\\ 
\hline
SA-PINN~\cite{Self-adaptivepinnjcp} &$1.27\pm 1.11\times 10^{-2}$\\
\hline
Hard Constraints~\cite{w712178} PINN &$7.03\pm 6.27\times 10^{-3}$ \\
\hline
MFF \cite{WANG202111}& $3.46\pm 0.51\times 10^{-3}$\\ 
\cline{1-2}
%Hard Constraints \cite{w712178}+MFF& $6.66\pm 3.49 \times 10^{-4}$\\ 
%\cline{1-2}
MMPINN-MFF (this work)&$6.56\pm 4.17\times 10^{-4}$\\ 
\cline{1-2}
MMPINN-INN (this work)&$2.13\pm 0.43\times 10^{-4}$\\ 
\cline{1-2}
\end{tabular}
\end{table}

\subsection{Klein–Gordon equation}\label{subsec:kgsection}

The Klein-Gordon equation  is a  relativistic counterpart of the Schrödinger equation, and has received considerable attention in its numerical and analytical aspects \cite{LIU20181}.
%and it is a relativistic form of Schrödinger equation used to describe particles with zero spin.
The example in this subsection is taken from Ref.~\cite{gradientpathologies}. 

Consider the 1D
Klein–Gordon equation as follows:

\begin{equation}\label{KG}
\begin{cases}
u_{tt} + \alpha u_{xx}+\beta u +\gamma u^k=f(x,t),& (x,t)\in \Omega \times (0,T),\\
u(x,0)=x,& x\in \Omega ,\\
u_t(x,0)=0,&  x\in \Omega,\\
u(x,t)=h(x,t),& (x,t)\in \partial \Omega \times [0,T] .\\
\end{cases}
\end{equation}
The model parameters are set with 
$\Omega=[0,1]$, $T=1$,  
$\alpha=-1$, $\beta=0$, $\gamma=1$, $k=3$. 
We use the fabricated solution 
\begin{equation}\label{kg exact solution}
u(x,t)=x \cos (a\pi t)+(xt)^3
\end{equation}
to assess the accuracy of the PINN methods.
$f(x,t)$ and $h(x,t)$ are derived from Eq.~\eqref{kg exact solution}. 

When the parameter $a$ in Eq.~\eqref{kg exact solution} is large, the conventional PINN method fails to produce an accurate prediction for this model. Wang et al. attribute the failure to the imbalanced gradient pathology. They separately propose a Learning Rate Annealing(LRA) algorithm  and an Improved Fully-connected Neural Network (IFNN), and combine them to improve the prediction accuracy~\cite{gradientpathologies}. 

With $a=5$ and $a=10$, we test the performance of different PINN methods.
The network architecture we use in this example contains 5 hidden layers, each layer with 150 neurons.
Two separate Fourier feature mappings are initialized with $\sigma_1=5$ and $\sigma_2=1$,
corresponding to $t$ and $x$, respectively.
The regularization parameters are set with $m=3$ and $n=3$. The batch sizes are set to $N_b=N_r=256$. In order to make a fair comparison with the methods proposed in Ref.~\cite{gradientpathologies}, the other hyperparameters are set according to the specifications of the article~\cite{gradientpathologies}.
The training of the neural networks uses the Adam optimizer to perform 40,000 iterations. The multi-level training strategy presented in section~\ref{Multi-level training strategy for multi-scale models} is not implemented in this example.

\begin{table}[h]
		\setlength{\abovecaptionskip}{0cm}
		\setlength{\belowcaptionskip}{0.2cm}
\caption{The $L_2$ relative errors of different methods for solving Eq.~\eqref{KG}.}
\label{kgKGlist}
\centering
\begin{tabular}{c|c|c}
\hline
\multirow{2}{*}{\makecell{Method}} & \multicolumn{2}{c}{ $\left\|\epsilon\right\|_2$} \\
\cline{2-3}
& $a=5$ & $a=10$ \\
\hline
Conventional PINN& $2.81\pm 0.76\times 10^{-2}$& $2.56\pm 0.57\times 10^{-1}$\\ 
\cline{1-3}
LRA with IFNN~\cite{gradientpathologies}&$1.42\pm 0.61\times 10^{-3}$& $9.55\pm 3.73\times 10^{-3}$\\ 
\cline{1-3}
MMPINN-MFF (this work)&$8.80\pm 2.03\times 10^{-4}$& $2.17\pm 0.57\times 10^{-3}$\\ 
\cline{1-3}
MMPINN-INN (this work)&$2.47\pm 0.50\times 10^{-4}$& $6.55\pm 1.18\times 10^{-4}$\\ 
\cline{1-3}
\end{tabular}
\end{table}

%change h to ht
\begin{figure}[ht]
    \raggedright % 图片靠左对齐
    \includegraphics[width=1\textwidth]{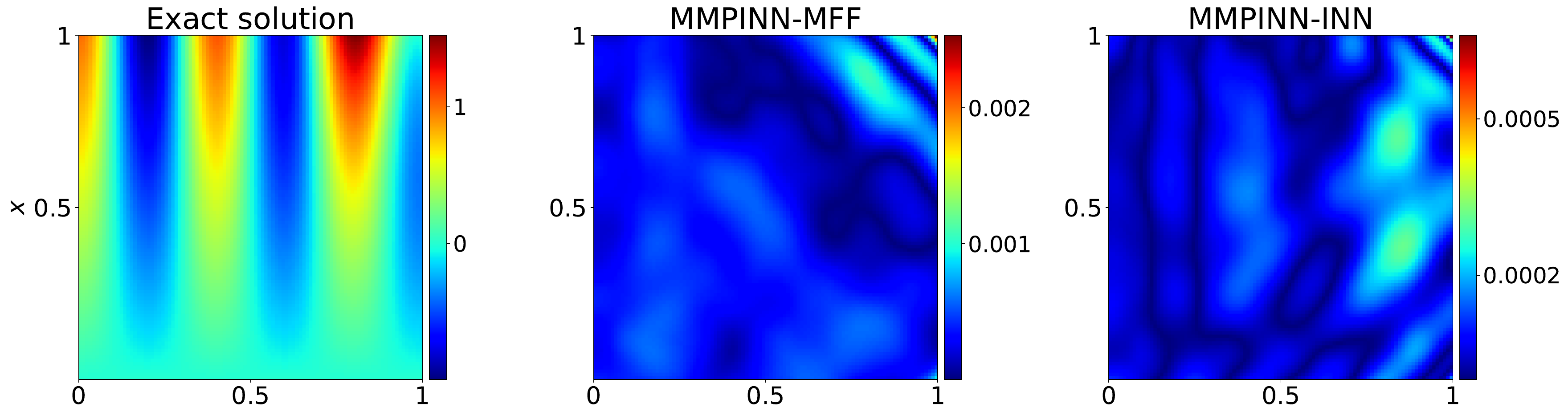}
     \includegraphics[width=1\textwidth]{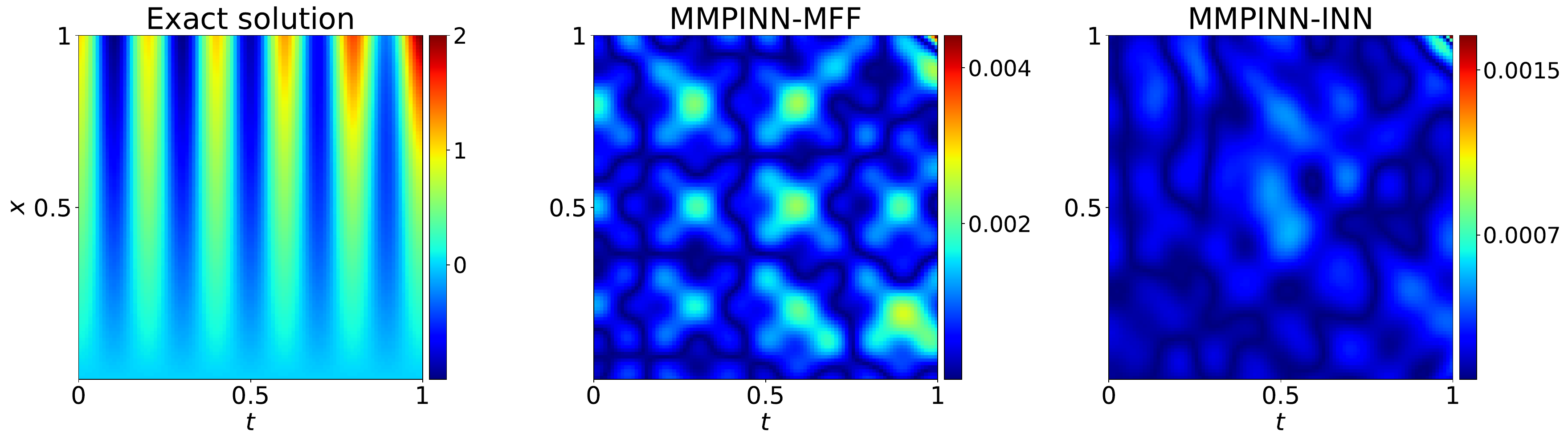} 
    \caption{The exact solution of Eq.~\eqref{KG} and the point-wise absolute errors of the MMPINN-MFF and MMPINN-INN methods
    (Line 1: $a=5$, Line 2: $a=10$).}
    \label{fig:kgeqution and point-wise error with mmpinn with ifnn}
\end{figure}

Table \ref{kgKGlist} shows that as the frequency increases, the accuracy of the method in Ref.~\cite{gradientpathologies} obviously decreases, while the MMPINN-INN method remains very good.
Figure \ref{fig:kgeqution and point-wise error with mmpinn with ifnn} provides a detailed comparison of the prediction accuracy, which also proves that the MMPINN-INN method is superior. 
The integrated neural network(INN) architecture plays a crucial role in improving the accuracy.

\subsection{A multi-frequency model }\label{subsec:multi-scale}

Both traditional numerical methods and deep learning methods are challenged by multi-frequency problems where the frequency varies significantly with position $x$ or time $t$. This example is designed to test the ability of our approaches to handle models with varying frequencies.
Consider a heat equation as follows:
\begin{equation}\label{multi-scale}
\begin{cases}
u_t=u_{xx}+f(x,t),& x\in(-1,1),t\in(0,1],\\
u(x,0)=0,& x\in(-1,1),\\
u(\pm 1,t)=\sin(2\pi t),&t\in(0,1].\\
\end{cases}
\end{equation}
The exact solution for this model is given by
\begin{equation}\label{exact multi scale solution}
u(x,t)=\sin\frac{20\pi t}{1+9x^2},
\end{equation}
$f(x,t)$ is derived from the exact solution. 

\begin{figure}[!h]
    \centering
    \includegraphics[width=1.0\textwidth]{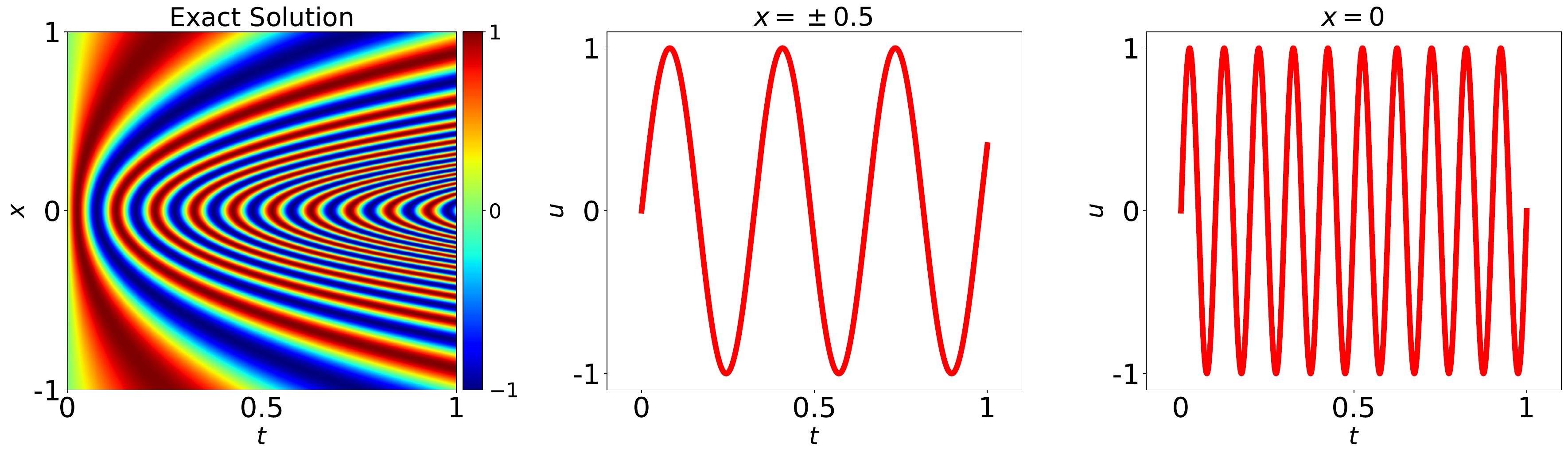} 
    \caption{The exact solution of Eq.~\eqref{multi-scale}.}
    \label{mfsolu}
\end{figure}

Figure \ref{mfsolu} shows the shape of the exact solution. The solution has a different frequency at different $x$.
In addition, this solution has a high frequency feature at the position $x=0$, which results in a large difference in order of magnitude between the supervised term and the residual term.
These two factors lead to the failure of training this model with the conventional PINN method. 
To obtain a valid prediction for this model, we use three improved PINN methods to train it.
A separate Fourier feature mapping is applied and initialized with $\sigma=10$.
The network architecture we use contains 4 hidden layers, each layer with 300 neurons. 
The numbers of training points are set with $N_0=1200$, $N_b=1200$ and $N_r=120000$. The regularization parameters are set with $m=3$, $n=3$.

\begin{table}[h]	\setlength{\abovecaptionskip}{0cm}
		\setlength{\belowcaptionskip}{0.2cm}
\caption{The $L_2$ relative errors of different methods for solving Eq.~\eqref{multi-scale}.}
\label{multi-scalelist}
\centering
\begin{tabular}{c|c} 
\hline 
Method&  $\left\|\epsilon\right\|_2$ \\
\hline
MFF \cite{WANG202111}& $1.03\pm 0.05\times 10^{-2}$\\ 
%MFF \cite{WANG202111}& $2.52\pm 0.42\times 10^{-3}$\\ 
%MFF \cite{WANG202111}& $1.59\pm 0.16\times 10^{-2}$\\ 
\cline{1-2}
Hard Constraints \cite{w712178} + MFF& $3.13\pm 0.77 \times 10^{-3}$\\ 
%MFF \cite{WANG202111}& $2.52\pm 0.42\times 10^{-3}$\\ 
%MFF \cite{WANG202111}& $1.59\pm 0.16\times 10^{-2}$\\ 
\cline{1-2}
%MMPINN-MFF (this work)&$3.88\pm 0.58\times 10^{-3}$\\ 
%MMPINN-MFF (this work)&$3.86\pm 0.56\times 10^{-4}$\\ 
MMPINN-MFF (this work)&$7.95\pm 1.41\times 10^{-4}$\\ 
\cline{1-2}
MMPINN-INN (this work)&$5.60\pm 1.79\times 10^{-4}$\\ 
\cline{1-2}
\end{tabular}
\end{table}

\begin{figure}[!h]
    \centering
    \includegraphics[width=1\textwidth]{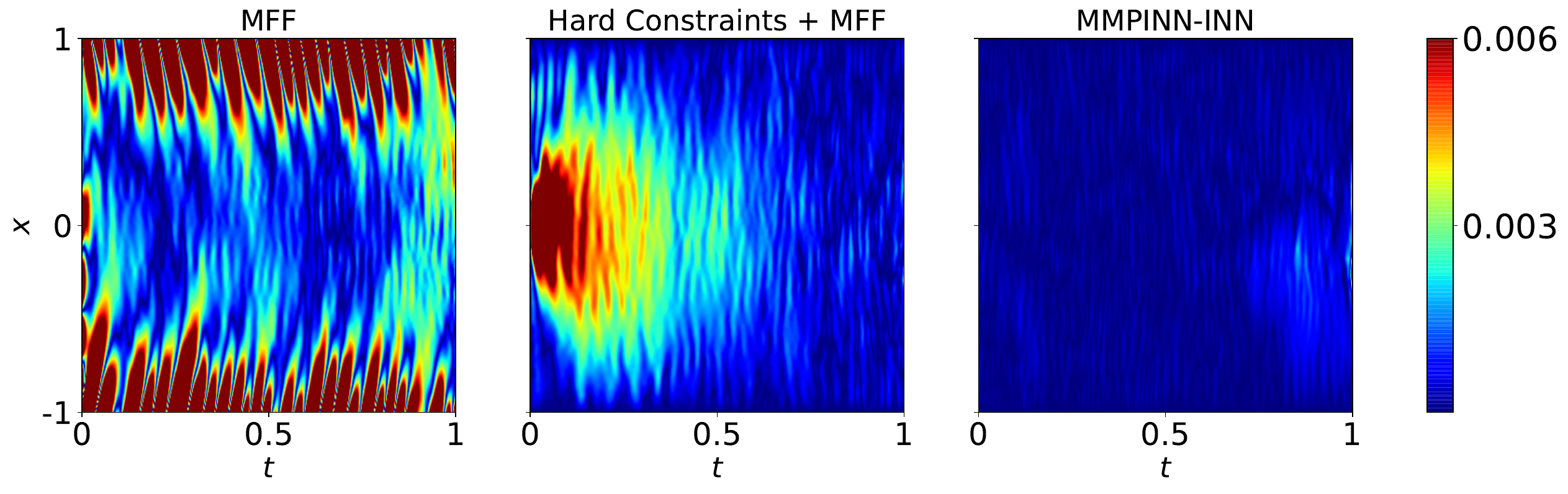} 
    \caption{The point-wise absolute errors of the MFF, Hard Constraints $+$ MFF and MMPINN-INN methods for solving Eq.~\eqref{multi-scale}.}
\label{tcgsmy}
\end{figure}

Table \ref{multi-scalelist} and Figure \ref{tcgsmy} show the results of the different methods. As can be seen from Figure \ref{tcgsmy}, the MFF method predicts the boundary conditions inaccurately, with large errors concentrated near the boundary, due to the lack of measures to reduce the difference in magnitude between the loss terms. Our methods are much more accurate than the MFF method, further demonstrating the perfection and necessity of combining MMPINN with INN.

\subsection{A problem with dramatic variations in different subdomains}\label{subsec:S44}

This example is designed to illustrate the importance of grouping regularization for the model whose residuals have large differences in different subdomains. 
Consider a Possion problem as follows: 
\begin{equation}\label{poissonE}
\begin{cases}
\Delta u(x,y)=f(x,y),&(x,y) \in \Omega =(0,1)\times(0,1),\\
\mathcal{B}(x,y)=g(x,y),&(x,y)\in \partial \Omega .\\
\end{cases}
\end{equation}
Suppose the exact solution is given by
\begin{equation}\label{exact poisson equation}
u(x,y)=1+(1000+y^2)e^{-\frac{(x-\frac{1}{2})^2}{2h^2}},
\end{equation}
where $h=0.02$.
$f(x,y)$ and $g(x,y)$ are derived from the exact solution.

\begin{figure}[h]
    \centering
\includegraphics[width=0.4\textwidth]{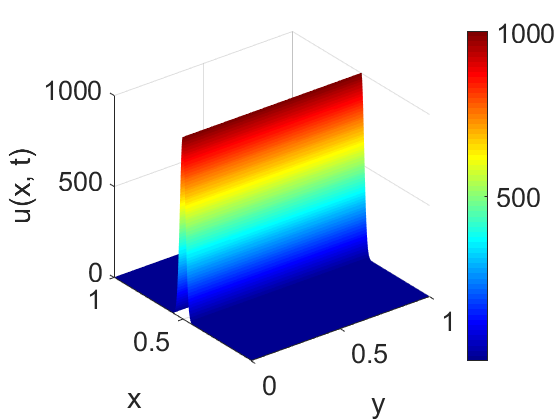}     
    \caption{The exact solution of Eq.~\eqref{poissonE}.}
    \label{fig:ESFor44}
\end{figure}

\begin{figure}[h]
    \centering
  \includegraphics[width=0.6\textwidth]{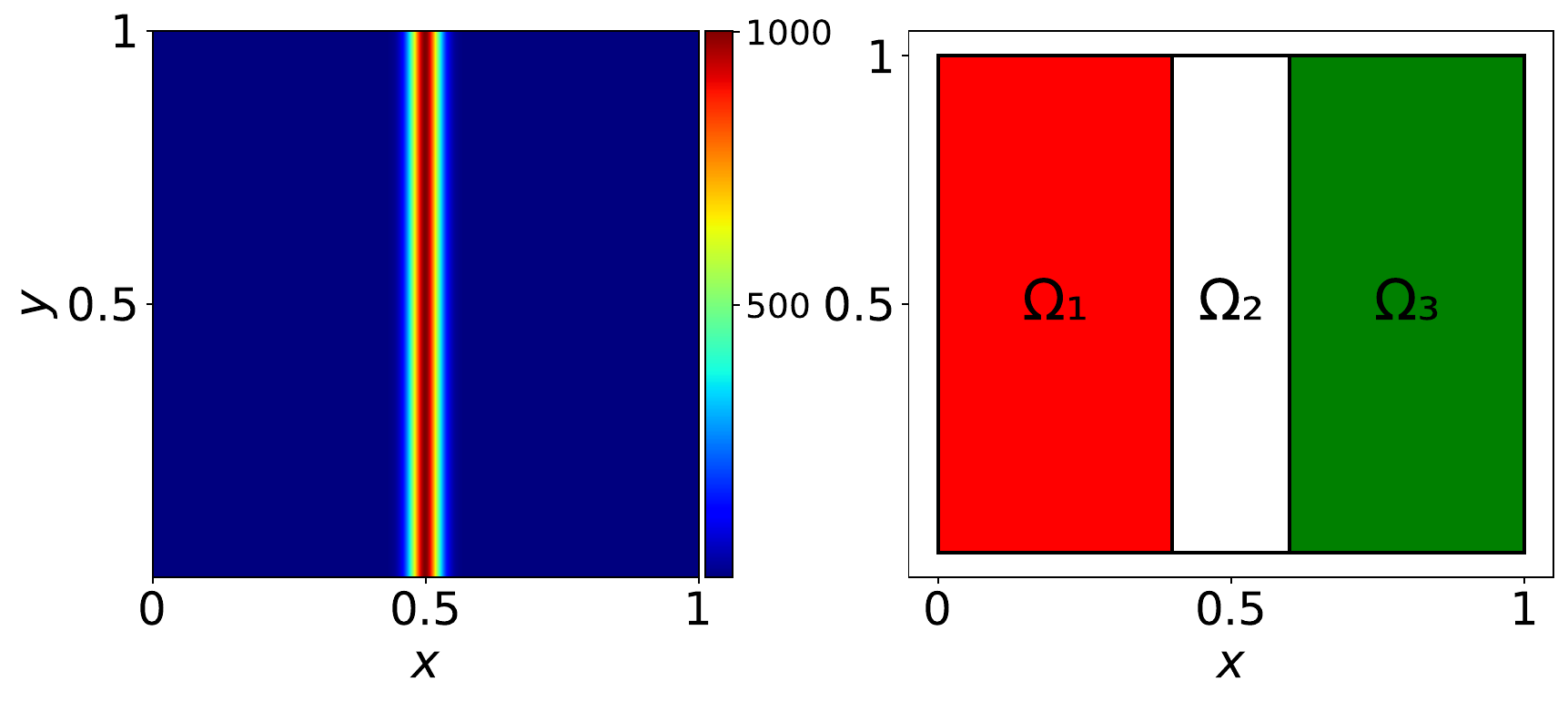} 
    \caption{Grouping regularization mode.}
    \label{fig:GRS}
\end{figure}

As shown in Figure \ref{fig:ESFor44}, the solution can be seen as a normal distribution function along the $x$-axis whose value reaches a maximum close to 1000 at $x$=0.5 and then decreases rapidly to a value close to 1 in a narrow interval.
Compared to other examples, this model has more obvious differences in different subdomains.
Therefore, to obtain an accurate prediction for this model, we have to use the grouping regularization strategy discussed in Section \ref{groupregstra}.
Based on the characteristics of the solution, we divide the whole domain $\Omega$ into three subdomains, as shown in Figure \ref{fig:GRS}. 
We refer to the MMPINN-DNN method using the grouping regularization strategy as \emph{MMPINN-DNN-GRS} in this benchmark.

The network architecture we use in this example contains 4 hidden layers, each layer with 50 neurons. 
The numbers of training points are set with $N_b=4800$ and $N_r=30000$. Note that for MMPINN-DNN-GRS, we set 10000 points in each subdomain.
The regularization parameters are set with $m=1$ and $n=3$ for the MMPINN-DNN method and $m=1$ and $n=(2,4,2)$ for the MMPINN-DNN-GRS method, which are determined by the ratios of the initial loss terms
$\mathcal{L}_s:\mathcal{L}_r=10^{4}:10^{11}$ and $\mathcal{L}_{\Omega_1}: \mathcal{L}_{\Omega_2}: \mathcal{L}_{\Omega_3} =10^{2}: 10^{11}: 10^{2} $.

Table \ref{PossioinEList} lists the results of the different methods. We see that it is difficult to obtain satisfactory predictions using conventional PINN methods. 
There are two factors that cause the PINN methods to perform poorly. The first factor is the large difference between the values of the supervised term and the residual term, which differ by about seven orders of magnitude.
The second factor is that the distribution of the residual term is extremely uneven, with the first and third parts differing by about nine orders of magnitude from the second part.

\begin{table}[h]	\setlength{\abovecaptionskip}{0cm}
		\setlength{\belowcaptionskip}{0.2cm}
\caption{The $L_2$ relative errors of different methods for solving Eq.~\eqref{poissonE}.}
\label{PossioinEList}
\centering
\begin{tabular}{c|c} 
\hline 
Method&  $\left\|\epsilon\right\|_2$ \\
\hline
conventional PINN& $1.12\pm 0.10\times 10^{0}$\\ 
\hline
conventional PINN ($10^6$ Adam)& $4.51\pm 2.81\times 10^{-2}$\\ 
\cline{1-2}
 MMPINN-DNN (this work)&$4.14\pm 3.29\times 10^{-3}$\\ 
\cline{1-2}
 MMPINN-DNN-GRS (this work)&$7.53\pm 1.64\times 10^{-4}$\\ 
\cline{1-2}
\end{tabular}
\end{table}

\begin{figure}[h]
    \centering
    \includegraphics[width=0.9\textwidth]{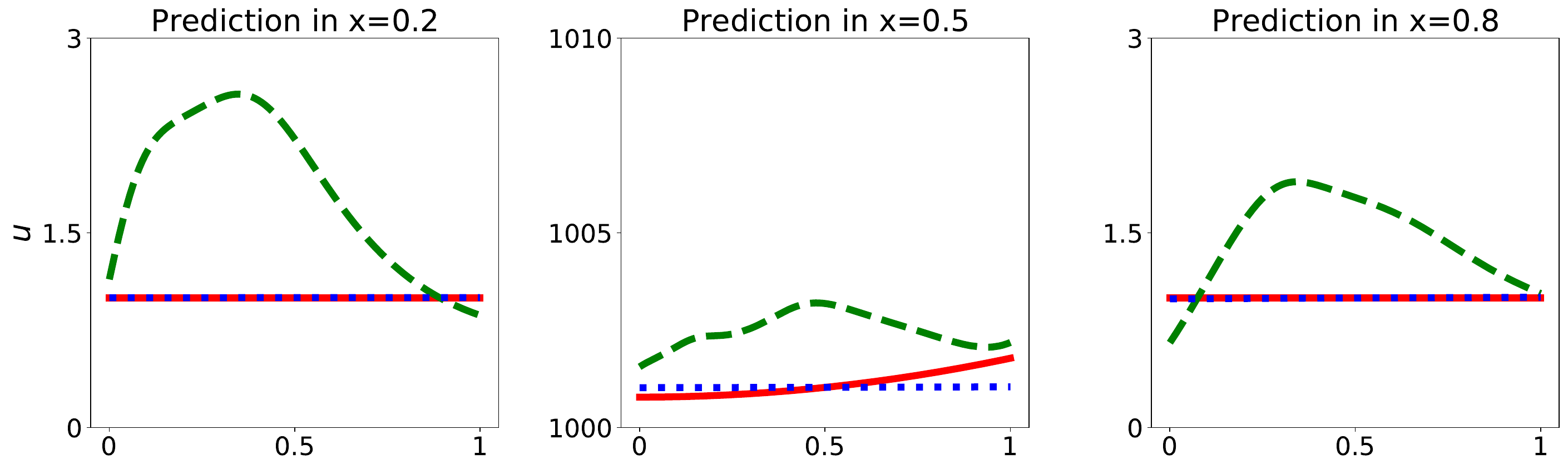} 
    \includegraphics[width=0.9\textwidth]{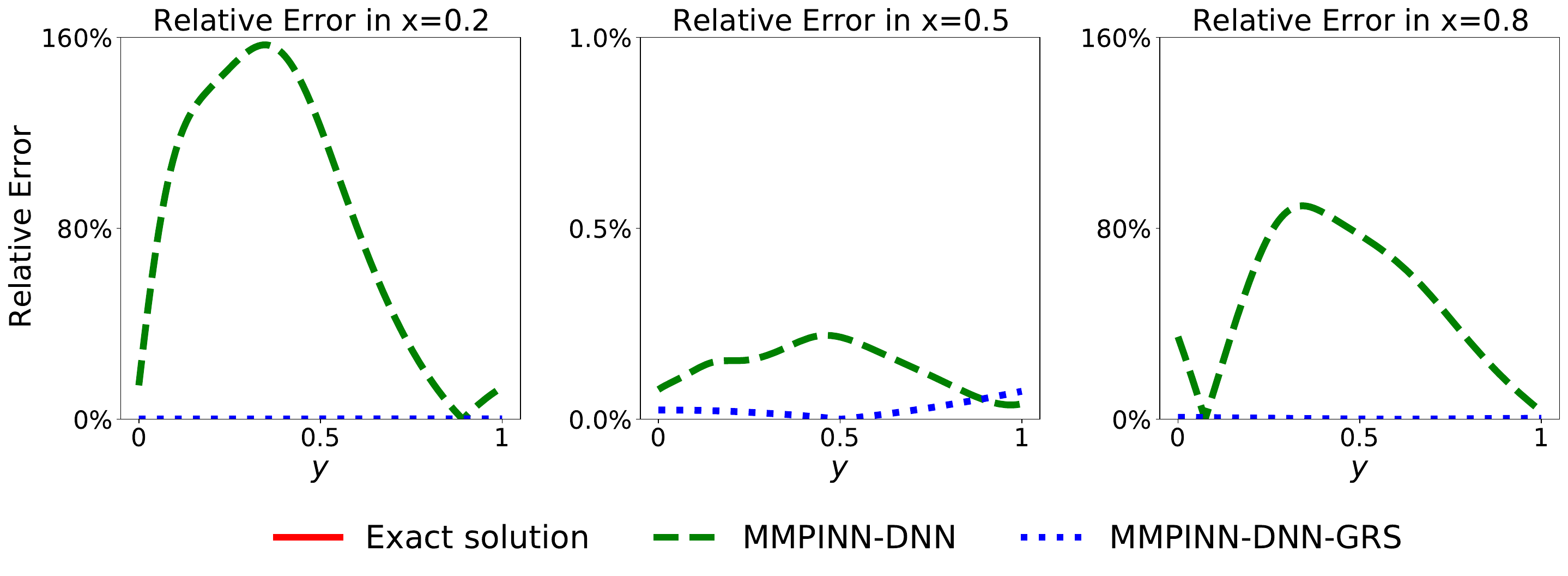} 
    \caption{Comparison of the MMPINN-DNN and MMPINN-DNN-GRS methods.}
    \label{fig:MMPINNVSMMPINGRS}
\end{figure}

The differences between the supervised term and the residual term could be balanced by MMPINN-DNN and Table \ref{PossioinEList} also shows that MMPINN-DNN can give more accurate prediction than PINNs.
However, because MMPINN-DNN does not address the issue of large differences between subdomains, so it cannot give satisfactory predictions for the subdomains with small residual terms, leading to large relative errors for $x = 0.2$ in $\Omega_1$ and $x = 0.8$ in $\Omega_3$, as shown in Figure \ref{fig:MMPINNVSMMPINGRS}.
The MMPINN-DNN-GRS method gives an accurate prediction for the entire domain, demonstrating the importance of the grouping regularization strategy.

\begin{figure}[!h]
    \centering
    \includegraphics[width=0.6\textwidth]{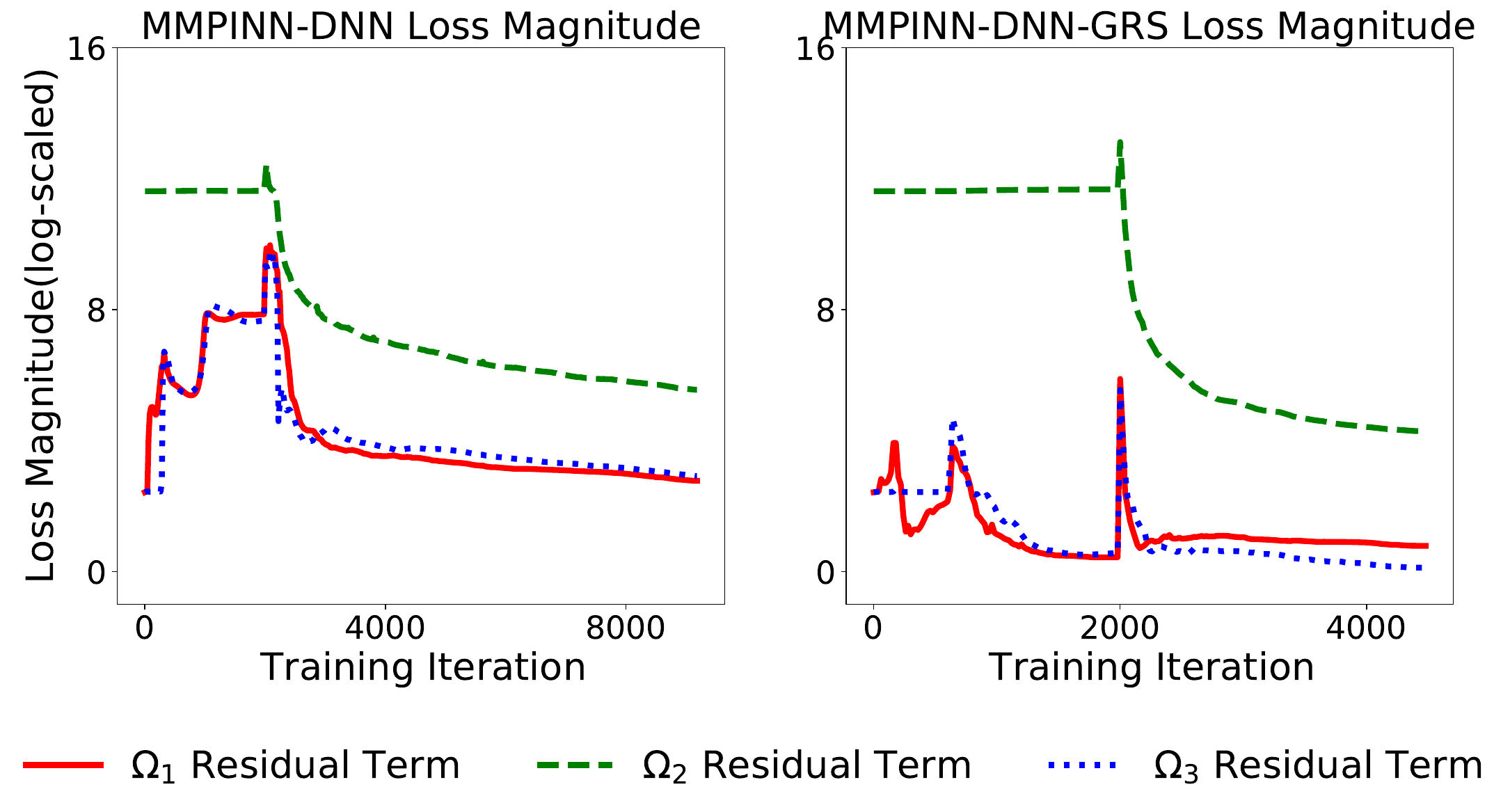} 
    \caption{Loss variation curves of MMPINN-DNN and MMPINN-DNN-GRS.
    %for solving Eq.~\eqref{poissonE}.
    }
    \label{fig:CurveforPossE}
\end{figure}

Figure~\ref{fig:CurveforPossE} shows that MMPINN-DNN-GRS further balances the difference of the residual terms between subdomains, which can cause the losses of all subdomains to decrease synchronously, resulting in more accurate predictions over the entire domain. 
In addition, the computational efficiency of MMPINN-DNN-GRS is higher than that of MMPINN-DNN. 
MMPINN-DNN-GRS requires about 4000 iterations to converge, while MMPINN-DNN requires nearly 10000 iterations.
Note that the jump at 2000 iterations is due to the optimizer switching from Adam to L-BFGS.

\section{Conclusion}
In this paper, we analyze and investigate the difficulties in solving multi-scale problems using the conventional PINN method.
For typical multi-scale models whose solutions have large gradients or high-frequency features, the supervised and residual terms of the loss function have large differences in magnitude. In this case, a pathology of the conventional PINN method is that there is a dominance of the PDE residuals in the optimization process. We eliminate this pathology by reconstructing the loss function using a novel regularization strategy for the loss terms, and propose an improved PNNN method denoted by MMPINN, which leads to a significant improvement not only in prediction accuracy but also in convergence speed. Compared with commonly used weighting methods, MMPINN makes the loss functions of physical neural networks more diverse, which opens a new door for tackling the magnitude difference in the loss function.
In addition, for the multi-scale problems with multi-frequency features, the common deep neural networks used by conventional PINNs have inherent deficiencies in dealing with this type of problem. To address this issue, we embed IFNN into MFF and develop an Integrated Neural Network(INN) architecture, which simultaneously mitigates spectral bias and gradient flow stiffness, and improves the computational accuracy of the multi-frequency problem.
Furthermore, by combining the above two approaches, we present the improved PINN framework, which is also called the MMPINN framework. 
The new methods derived from the MMPINN framework handle the multi-scale problems very well, and their performance outperforms many similar methods.

%Through in-depth investigation, we further develop MFF architecture \cite{WANG202111} 
%It is worth noting that MMPINN is not limited to PINNs, but can be generalized to other multitask learning, such as Deep Galerkin Method \cite{SIRIGNANO20181339}.

Although we achieve some interesting results in this paper, we acknowledge that the MMPINN framework is still in its early stages.  In order to advance the MMPINN framework, we focus on the following issues in our future work:  
How to adaptively select the regularization parameters $(m,n)$? How to choose an appropriate neural network architecture without prior knowledge? Since the INN architecture requires much more computational resources, how to prune the neurons while maintaining relatively high accuracy? Can the grouping strategy be automated during training?
%One main limitation is a lack of rigorous mathematical proof, although  we have verified the effectiveness of the method under challenging settings.
%What is the relationship between MMPINN and stiffness in gradient flow dynamics?
%Can we combine MMPINN with more state-of-the-art network architectures?

%We believe that answering these questions will shed light on the development of the MMPINN framework for solving more realistic multi-scale problems.

\section*{Acknowledgements}
The work is supported by the National Key R$\&$D Program of China under Grant No.2022YFA1004500, the National Science Foundation of China under Grant No.12271055 and No.12171048.
We thank the authors of the papers~\cite{WANG202111,gradientpathologies} for making their codes public, which helps to compare different methods.
%The authors thank Dr.~Miao for providing the reference solution of Sect.~\ref{subsec:S43}.

\bibliography{mybibfile}

\end{document}